# A Robust Multimodal Remote Sensing Image Registration Method and System Using Steerable Filters with First- and Second-order Gradients

Yuanxin Ye[a, b], Bai Zhu[a, b], Tengfeng Tang[a,b], Chao Yang[a,b], Qizhi Xu[c], Guo Zhang[d, *]

[a]Faculty of Geosciences and Environmental Engineering, Southwest Jiaotong University, Chengdu 610031, China

[b]State-province Joint Engineering Laboratory of Spatial Information Technology for High-speed Railway Safety, Southwest Jiaotong University, Chengdu 610031, China

[c]School of Mechatronical Engineering, Beijing Institute of Technology, Beijing 100081, China

[d]State Key Laboratory of Information Engineering in Surveying, Mapping and Remote Sensing, Hubei, Wuhan 430079, China

**ABSTRACT:**

Co-registration of multimodal remote sensing (RS) images (e.g., optical, infrared, LiDAR, and SAR) is still an ongoing challenge because of nonlinear radiometric differences (NRD) and significant geometric distortions (e.g., scale and rotation changes) between these images. In this paper, a robust matching method based on the Steerable filters is proposed consisting of two critical steps. First, to address severe NRD, a novel structural descriptor named the Steerable Filters of first- and second-Order Channels (SFOC) is constructed, which combines the first- and second-order gradient information by using the steerable filters with a multi-scale strategy to depict more discriminative structure features of images. Then, a fast similarity measure is established called Fast Normalized Cross-Correlation (Fast-NCC$_{SFOC}$), which employs the Fast Fourier Transform (FFT) technique and the integral image to improve the matching efficiency. Furthermore, to achieve reliable registration performance, a coarse-to-fine multimodal registration system is designed consisting of two pivotal modules. The local coarse registration is first conducted by involving both detection of interest points (IPs) and local geometric correction, which effectively utilizes the prior georeferencing information of RS images to address global geometric distortions. In the fine registration stage, the proposed SFOC is used to resist significant NRD, and to detect control points (CPs) between multimodal images by a template matching scheme. The performance of the proposed matching method has been evaluated with many different kinds of multimodal RS images. The results show its superior matching performance compared with the state-of-the-art methods. Moreover, the designed registration system also outperforms the popular commercial software (e.g., ENVI, ERDAS, and PCI) in both registration accuracy and computational efficiency. Our system is available at https://github.com/yeyuanxin110/SFOC-Multimodal_Remote_Sensing_Image_Registration_System.

---

[*] Corresponding author: Guo Zhang, guozhang@whu.edu.cn

**KEYWORDS:** Multimodal images, SFOC, Fast-NCC$_{SFOC}$, integral feature images, Registration system

## 1. INTRODUCTION

Nowadays, image registration drives extensive application in the fields of remote sensing, computer vision, and medical imaging. In general, image registration is a prerequisite step for remote sensing (RS) image processing and analysis applications, such as image fusion (Stathaki, 2011), change detection (Seydi et al., 2020), and environmental monitoring (Behling et al., 2014). Moreover, the effectiveness of these subsequent applications is often directly affected by registration accuracy, even if misregistration is only within a pixel range. The key to RS image registration is to find an evenly distributed and high-precision set of control points (CPs) as much as possible, and then to calculate the optimal geometric transformation model. The rapid and explosive growth of RS image datasets (e.g., optical, infrared, SAR, and LiDAR) promotes the development of the aerospace industry. However, these RS images are usually captured by either different sensors from different perspectives or the same sensor in different periods (Zitova and Flusser, 2003). These factors have brought a great challenge for achieving precise co-registration, and it is difficult to develop a fully universal method to cope with all registration cases. Any kind of image registration algorithm needs to consider the imaging principle, radiometric and geometric distortions, noise interference, clouds occlusions, and so on.

Generally, the georeferencing of RS image data can be divided into two different types: images with direct geo-referencing have been geometrically corrected (i.e., Level 2 data), and images with the rational polynomial coefficients (RPCs) without geometric correction (i.e., Level 1 data). Since the RPCs can be supplied by commercial satellite image vendors instead of the rigorous sensor model to conceal all the technical details relating to cameras and satellite orbits (Zhang and Zhu, 2008), it has been widely used for conducting the georeferencing of high-resolution satellite imagery (Shen et al., 2017). However, the RPCs are usually biased that caused by the inaccurate measurement of the satellite ephemeris and instrument calibration, which results in the georeferencing images having an offset typically ranging from several pixels to dozens of pixels in the image space (Jiang et al., 2015; Xiang et al., 2021). Fig. 1 exemplarily shows two pairs of multimodal images with the direct geo-referencing, which include a pair of Google and GaoFen-2 Panchromatic images and a pair of Google and Sentinel-1 SAR images. It can be clearly observed that the implementation of georeferencing only can address the obvious global geometric differences (e.g., rotation and scale changes).

Although the georeferencing implementation just eliminates global geometric distortions, it can provide assistance for subsequent fine registration. It is evident from Fig. 1 that the most serious challenge for fine registration is nonlinear radiometric differences (NRD) between multimodal images. Moreover, the interference of strong speckle noise is very serious on the SAR images, which is also an inevitable problem for the registration of SAR and other types of images. These challenges make it difficult to detect precise CPs even by visual inspection. Therefore, this paper will make use of the georeferencing information to handle global geometric differences, and then develop a robust matching method to resist NRD and noise interference, realizing the fast and robust registration for multimodal RS images.

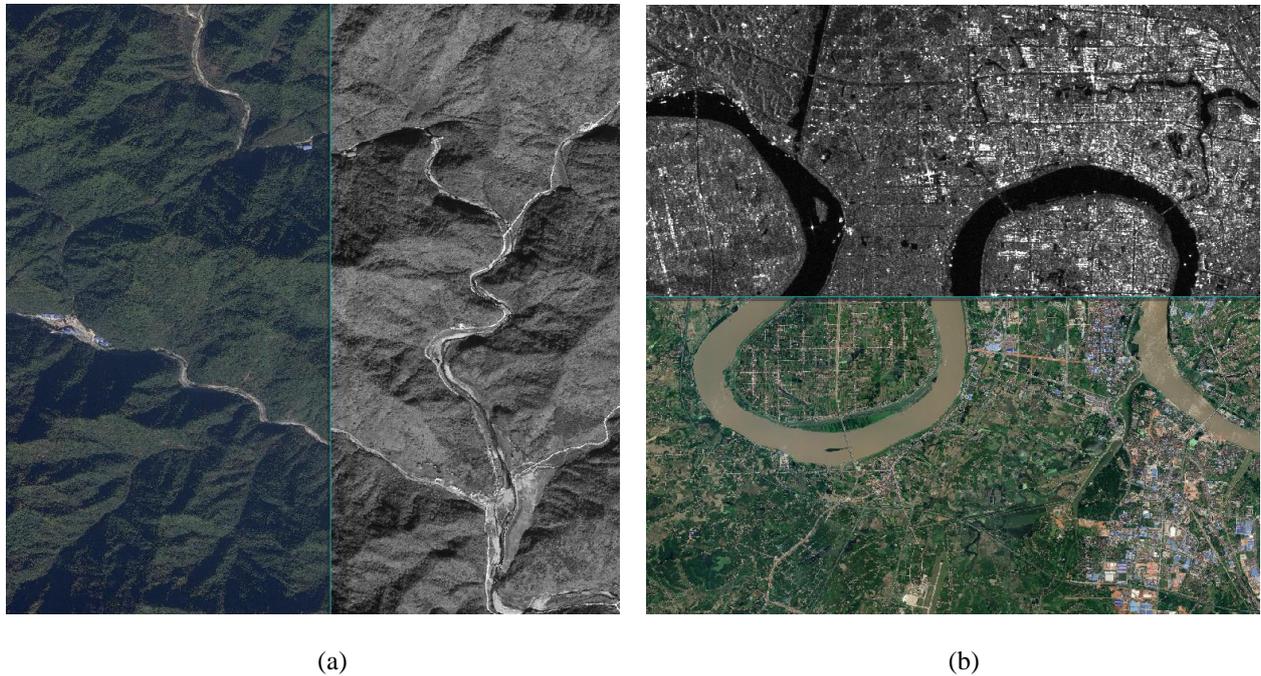

(a) (b)

Fig. 1. Example of multimodal images with direct georeferencing. (a) Google (Left) and Panchromatic (Right) images. (b) Google (Nether) and SAR (Upper) images.

To date, numerous efforts have been made to overcome these above challenges and improve the performance of multimodal RS image registration. These methods can be commonly classified into three categories with the taxonomy of intensity-based methods (IBM), feature-based methods (FBM), and learning-based methods (LBM) (Jiang et al., 2021).

IBM evaluates the similarity of intensity information by using a template matching strategy, which relies on the selection of similarity measures that play a pivotal role in this process. According to the different image representation domains, IBM can usually be divided into spatial domain and frequency domain. The most common similarity measures consist of three types in the spatial domain: the sum of squared differences (SSD) (Zitova and Flusser, 2003), the normalized cross-correlation (NCC) (Yun-hui, 2013), and the mutual information (MI) (Chen et al., 2003). These measures have been used extensively in image registration. Nonetheless, SSD and NCC are sensitive to nonlinear radiometric differences (NRD) that generally exist in different kinds of multimodal RS images (Uss et al., 2016). Although MI has been testified to be effective for resisting NRD, MI is clumsy and time-consuming because it must compute the joint histogram based on statistical similarity (Suri and Reinartz, 2009). To improve the computational efficiency, the phase correlation (Reddy and Chatterji, 1996) is expanded to align images by utilizing the Fast Fourier Transform (FFT) in the frequency domain (Xiang et al., 2020). Tong et al. (2015) presented a subpixel phase correlation method using singular value decomposition (SVD) and the unified random sample consensus (RANSAC) (Fischler and Bolles, 1981) to improve the robustness of multispectral image registration. Wan et al. (2019) designed a stepwise algorithm, called least-squares fitting-based phase correlation (SLSF-PC), to perform illumination-insensitive image matching. Although the phase correlation has illumination invariance, it is not enough robust to NRD and noise that exist in multimodal images, especially in optical and SAR images. Therefore, these disadvantages limit the application of IBM in the multimodal registration field.

FBM differs from IBM to comprise the remarkable and invariant features (e.g., point features, line features, and region features), which evaluates the similarity of these significant features rather than intensity information to achieve registration. Such methods generally consist of common feature extraction and feature matching, with the most common method to be Scale Invariant Feature Transform (SIFT) (Lowe, 2004) and its variants, such as SAR-SIFT (Dellinger et al., 2014), adaptive binning SIFT (Sedaghat and Ebadi, 2015), and OS-SIFT (Xiang et al., 2018). The above algorithms take advantage of these invariant features to resist geometric distortions, but it is difficult to extract a large number of uniform and stable features from multimodal images with significant NRD, which has its limitations in multimodal registration application. To tackle these problems, a growing number of novel and valid descriptors have been designed based on structural and shape features, which are inspired by multimodal natural image matching (Shechtman and Irani, 2007) and multimodal medical image registration (Heinrich et al., 2012). Given the advantages of phase congruency in image perception (Morrone and Owens, 1987), Ye et al. (2017) designed a histogram of orientated phase congruency (HOPC) descriptor by integrating the phase congruency with an orientation histogram to extract distinct structural features. Furthermore, Cui and Zhong, (2018) constructed a multi-scale phase consistency (MSPC) taking advantage of the Log-Gabor odd-symmetric wavelet, then the Euclidean distance of the MSPC descriptor was used as a similarity measure to detect correspondences. Similarly, a phase congruency structural descriptor (PCSD) was established using the phase congruency structure image by (Fan et al., 2018), which extracts structure information by employing the Log-Gabor filter responses over different orientations and scales. With the superiority of the phase congruency in resisting NRD, Li et al. (2019) proposed a radiation insensitive feature based on phase congruency and a maximum index map, named radiation-variation insensitive feature transform (RIFT), which has been successfully validated on different types of multimodal images. Although these phase-congruency-based methods have been proven to improve the performance of multimodal registration, they required the amplitude and orientation of phase congruency, leading to complicated calculation and time-consuming processes.

As deep learning has shown superior performance in image matching in the field of computer vision (Dusmanu et al., 2019), LBM has also been introduced into the remote sensing image registration field. Ye et al. (2018) designed a novel method for multispectral image registration using the combination of middle- or high-level features extracted by a convolutional neural network (CNN) and the SIFT descriptor, which can overcome the defect of the original SIFT and improve the performance of registration. Wang et al. (2018b) constructed an end-to-end deep learning architecture that directly learns the mapping function between the sensed and reference image patch-pairs and their corresponding matching labels, where a transfer learning based on self-learning was used to accelerate the training process. Ma et al. (2019) introduced a coarse-to-fine registration method based on CNN and local features, where the CNN features were first detected by using VGG-16 to achieve deep pyramid feature representation. Subsequently, the feature matching and transformation estimation were further refined by combining the deep CNN features and handcrafted local features. Similarly, Zhang et al. (2019) evaluated the similarity score related to the learned common features extracted by their designed Siamese fully convolutional network (SFcNet), in this way, outlier can be removed and successful registration of a variety of multimodal images can be realized. Hughes et al. (2020) proposed a three-step matching framework consisting of a goodness network, multi-scale matching network, and outlier reduction network to realize the fully automated and multi-scale SAR and optical image registration. Zhou et al. (2021) first adopted a shallow pseudo-Siamese network with a small number of model parameters to produce the multiscale convolutional gradient features (MCGFs), then the refined MCGF was used for improving the matching capacity of SAR and optical images. Cui et al. (2021) employed an attention mechanism and the spatial pyramid aggregated pooling to construct a novel network, namely MAP-Net with robustness to NRD and geometric distortions, which makes the designed key features containing the high-level semantic information, and this is beneficial to register multimodal images.

Although current LBMs have achieved remarkable progress, their disadvantages are also quite significant. The main drawback is that LBM usually requires a large amount of training and labeled data, which will greatly affect the practical application of image registration. Due to the hyperparameters of training a deep network is generally far more than a human-defined feature extractor, the training efficiency is greatly related to the basic configuration of computer infrastructure. LBM's superiority only is brought into play in multimodal image registration based on high-performance computer infrastructures, which is another disadvantage to limit its widespread use.

Moreover, mainstream commercial softwares (e.g., ENVI, ERDAS, and PCI) have basic registration function modules in which the traditional registration measures (e.g., NCC and MI) are still adopted for automatic multimodal image registration. As mentioned above, these traditional measures are difficult to achieve high precision registration of multimodal images that exist significant NRD and geometric distortions. Therefore, there is a great demand for developing a robust and automatic registration system of multimodal images in engineering practice.

In order to meet these above registration requirements, this paper first proposes a fast and robust matching method that is composed mainly of two essential components, then further designs an efficient registration system that involves a coarse-to-fine process to cope with various multimodal remote sensing images. During the matching stage, we first construct a novel and discriminative descriptor, called the Steerable Filters of first- and second-Order Channels (SFOC), through combining the first-order gradients with the second-order gradients by using the steerable filters, which is utilized to address significant NRD between multimodal images. Then, we establish a fast similarity measure, namely Fast Normalized Cross-Correlation (Fast-NCC$_{SFOC}$), by improving the traditional NCC using the Fast Fourier Transform (FFT) technique and integral images, which is employed to accelerate the matching process. During the implementation of the registration system, a local coarse registration is performed by carrying out a Features from Accelerated Segment Test (FAST) operator (Rosten and Drummond, 2006) with a partitioning strategy to detect uniformly distributed IPs, and whereafter designing a local geometric correction based on the Rational Function Model (RFM) to eliminate global geometric distortions. In the subsequent fine registration stage, the designed system detects control points (CPs) by employing the proposed matching method and removing larger outliers by the RANSAC approach.

In general, the main contributions of this paper are listed as follows: (1) a robust matching method consisting of two main components, a novel structure descriptor (SFOC) that combines the first-order gradients and second-order gradients by using the steerable filters with a multi-scale strategy to depict multidirectional and multiscale structure characteristics, and a fast similarity measure (Fast-NCC$_{SFOC}$) using the FFT technique and integral images to improve the matching efficiency. (2) an efficient registration system involving local coarse registration and fine registration. The local coarse registration makes full use of the prior georeferencing information of remote sensing images based on the RFM model to provide reliable spatial geometry constraints, which facilitates the subsequent fine registration.

This paper is a further extension of our previous work (Ye et al., 2021a), specifically as follows: (1) a novel descriptor named SFOC that is presented to depict more discriminative structure features by using the steerable filters with first- and second-order gradients. (2) a fast similarity measure called Fast-NCC$_{SFOC}$ that is proposed to detect CPs by utilizing the FFT technique and integral images. (3) a more thorough evaluation of both the proposed matching method and the designed system using more multimodal RS images.

The rest of this paper is organized as follows. Section 2 elaborates the pivotal components of the proposed matching method, including a discriminative descriptor and a fast similarity measure. In Section 3, the designed coarse-to-fine registration system is then introduced in detail. Subsequently, Section 4 evaluates the performance of the proposed matching method by using diverse multimodal datasets. In Section 5, the registration results of the designed system are discussed and analyzed. Finally, the conclusions are presented in Section 6.

## 2. PROPOSED MATCHING METHOD

In this section, a robust and fast matching method is proposed that consists of two key components. The two components are a novel structural feature descriptor and a fast similarity measure, respectively. The detailed instruction will be provided in the following sections.

### 2.1 Construction of structural feature descriptor

As mentioned above, the structural features exhibit excellent performance in multimodal image registration, and the phase-congruency-based methods have high time complexity despite their robustness to NRD (Wang et al., 2020; Li et al., 2019; Cui and Zhong, 2018; Ye et al., 2017). Recently, a number of descriptors based on multi-orientated gradient information to depict structural features have also proved to be robust to NRD, among which the channel features of orientated gradients (CFOG) (Ye et al., 2019), the angle-weighted oriented gradient (AWOG) (Fan et al., 2021), and the multi-Scale and multi-Directional Features of odd Gabor (SDFG) (Zhu et al., 2021) are the most representative ones. Specifically, CFOG first explored the pixel-wise feature representation based on the dense orientated gradient features to efficiently capture structure features of images and achieve fast and robust matching for multimodal images. However, CFOG generated multi-direction gradient features by the interpolation of the horizontal and vertical gradients, which results in reducing the distinguishability of the features. AWOG proposed an angle-weighted strategy to allocate the gradient values into the two most related orientations rather than interpolating to arbitrary direction as in the case of CFOG, and in this way, the distinctiveness of its feature vectors can be enhanced. The SDFG descriptor further increased a multi-scale strategy with multi-direction description using odd Gabor functions to extract the multiscale and multi-directional structural features, which can also further improve the discrimination of features.

Although the CFOG, AWOG, and SDFG descriptors have been successfully used for multimodal image matching, the construction of gradient channels for CFOG is calculated by simple pixel differences, which are very sensitive to noises. While the horizontal and vertical gradients of AWOG are calculated by the Sobel operator that simply comprises the first-order x-derivative and y-derivative operators. Meanwhile, the multi-scale information is deficient due to both the CFOG and AWOG neglecting the local inter-pixel relationships of images. Despite SDFG integrating the multi-scale information for feature description, it is similar to CFOG and AWOG that only make use of the first-order gradients, which results in a lack of local shape attributes in terms of curvature that exploited by the second-order gradients (Huang et al., 2014).

Therefore, our work is strongly driven and inspired by the three similar structural feature descriptors (i.e., CFOG, AWOG, and SDFG) and to some extent also by the phase-congruency-based methods. The main purpose of the work in this paper is to construct a descriptor that captures distinctive structural features of images from different modalities as robust as possible. In this subsection, a novel structural descriptor is specifically proposed, named the Steerable Filters of first- and second-Order Channels (SFOC), for depicting multi-directional and multi-scale structural characteristics of multimodal images. Unlike the phase-congruency-based methods and the SDFG descriptor, the SFOC extracts multidirectional and multiscale structure characteristics by taking advantage of the steerable filters with a multi-scale strategy and the dilated Gaussian convolution with different dilated rates. In such a way, the loss of local spatial information caused by the gradient difference of CFOG and the Sobel operator of AWOG can be avoided, and the anti-interference for noise can be improved.

If a function can be represented as a linear combination of rotated versions of itself, it is considered "steerable". The steerable filters refer to a class of arbitrary orientation filters that can be synthesized into a linear combination of base filters (Freeman and Adelson, 1991). Therefore, the steerable filters can adjust different angles to realize the adaptive control of the filters, with linear, multi-direction, and multi-scale characteristics, so as to provide more details in the image information of direction and edges, and have a wide range of applications in the field of contour extraction (Kochner et al., 1998), feature detection (Jacob and Unser, 2004), image denoising (Rabbani, 2009) and image enhancement (Zheng et al., 2019). More details regarding the steerable filters that are exploited for constructing the proposed descriptor are provided below.

The higher-order directional derivatives of the Gaussian function have been proved to be steerable, among which the simplest steerable filter is the first-order Gaussian derivative. The Gaussian function $G(x)$ in two-dimensional space is shown in the following equation:

$$G(x) = \frac{1}{2\pi\sigma^2} e^{\frac{-(x^2+y^2)}{2\sigma^2}} \tag{1}$$

Where (x, y) are Cartesian coordinates, $\sigma$ represents the variance of Gaussian function. Let $G_n$ be the $n$th derivative of the $G(x)$ in the x-direction, and $\theta$ represents the rotation of any function concerning the origin. The first-order $x$ Gaussian derivative is expressed as follows:

$$G_{1,\sigma}^{0°} = \frac{\partial G}{\partial x} = (-\frac{1}{2\pi\sigma^4})xe^{\frac{-(x^2+y^2)}{2\sigma^2}} \tag{2}$$

If the same function $G(x)$ is rotated 90°, the following equation can be obtained:

$$G_{1,\sigma}^{90°} = \frac{\partial G}{\partial y} = (-\frac{1}{2\pi\sigma^4})ye^{\frac{-(x^2+y^2)}{2\sigma^2}} \tag{3}$$

The first-order steerable $G_1$ filter at arbitrary orientation $\theta$ can be synthetized by making use of a linear combination of $G_{1,\sigma}^{0°}$ and $G_{1,\sigma}^{90°}$:

$$G_{1,\sigma}^{\theta} = \cos(\theta)G_{1,\sigma}^{0°} + \sin(\theta)G_{1,\sigma}^{90°} \tag{4}$$

Where $G_{1,\sigma}^{0°}$ and $G_{1,\sigma}^{90°}$ are regarded as the basis filters of $G_{1,\sigma}^{\theta}$ filter because all the sets of $G_{1,\sigma}^{\theta}$ can be combined by them.

Moreover, recent studies (Morgan, 2011; Wallis and Georgeson, 2012) have shown that many local feature descriptors based on the first-order gradient information, such as SIFT, Histogram of Oriented Gradient (HOG) (Dalal and Triggs, 2005), DAISY (Tola et al., 2009), are far from accurate in capturing visual features of human perception. Since the first- and second-order gradients are related to different geometric and structural features of images, the second-order gradient has better performance in describing detailed information than the first-order gradient. Hence, a more discriminative structure feature of the image can be depicted and reinforced when they are used in combination (Huang et al., 2014). In addition to the first-order steerable $G_1$ filter, therefore, the second-order steerable $G_2$ filter is also used in subsequent descriptor construction. Similar to the steerable $G_1$ filter, the second-order Gaussian steerable filter $G_2$ (Freeman and Adelson, 1991; Liu et al., 2002) is defined as follows:

$$\begin{cases} G_{2,\sigma}^{\theta} = k_1(\theta)G_{2,\sigma}^{0°} + k_2(\theta)G_{2,\sigma}^{60°} + k_3(\theta)G_{2,\sigma}^{120°} \\ G_{2,\sigma}^{0°} = G_{xx} = (-\frac{1}{2\pi\sigma^4})(1-\frac{x^2}{\sigma^2})e^{\frac{-(x^2+y^2)}{2\sigma^2}} \\ G_{2,\sigma}^{90°} = G_{yy} = (-\frac{1}{2\pi\sigma^4})(1-\frac{y^2}{\sigma^2})e^{\frac{-(x^2+y^2)}{2\sigma^2}} \\ G_{xy} = (\frac{xy}{2\pi\sigma^6})e^{\frac{-(x^2+y^2)}{2\sigma^2}}, G_2^{60°} = G_{yy} - G_{xy}, G_2^{120°} = G_{yy} + G_{xy} \\ k_j(\theta) = \frac{1}{3}[1+2\cos(2(\theta-\theta^j))], \theta_1 = 0°, \theta_2 = 60°, \theta_3 = 120° \end{cases} \quad (5)$$

Where $G_{2,\sigma}^{0}$, $G_{2,\sigma}^{60}$, and $G_{2,\sigma}^{120}$ are regarded as the basis filters of $G_{2,\sigma}^{\theta}$ filter because all the sets of $G_{2,\sigma}^{\theta}$ can be combined by them, $k_j(\theta)$ represents the corresponding interpolation functions.

Formally, the construction of the proposed SFOC descriptor mainly consists of three key components: (1) the construction of the first-order steerable channels with multi-scale strategy, (2) the construction of the second-order steerable channels, and (3) dilated Gaussian convolution and normalization. Fig. 2 demonstrates the construction flowchart of the proposed SFOC descriptor and more details of which are specified as follows.

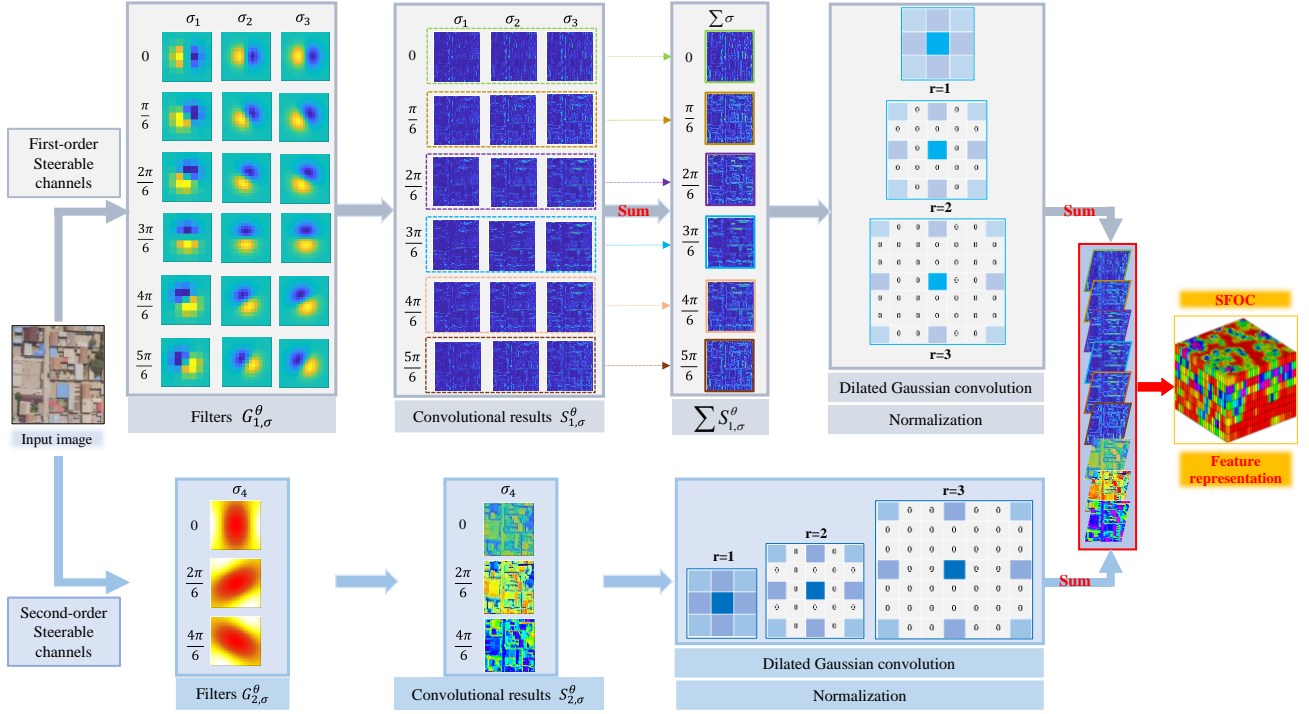

Fig. 2. Construction flowchart of the proposed SFOC descriptor

The construction of SFOC is divided into two critical channels: the first-order steerable channels and the second-order steerable channels. Since the convolution operation is a linear operator, thus the first-order steerable channels of an image $I(x, y)$ at an arbitrary orientation $\theta$ can be computed by convoluting the image with $G_{1,\sigma}^{0°}$ and $G_{1,\sigma}^{90°}$. In the proposed descriptor, the establishment of first-order channels is composed of six directions: $0, \frac{\pi}{6}, \frac{2\pi}{6}, \frac{3\pi}{6}, \frac{4\pi}{6}, \frac{5\pi}{6}$. Meanwhile, the multi-scale strategy with different Gaussian standard deviations (STD) is embedded to further reinforce the descriptive completeness of local structure features with the purpose of increasing the discrimination. The specific calculation process is as follows:

$$\begin{cases} S_{1,\sigma}^{0°} = G_{1,\sigma}^{0°} * I(x, y) \\ S_{1,\sigma}^{90°} = G_{1,\sigma}^{90°} * I(x, y) \\ S_{1,\sigma}^{\theta} = \cos(\theta) S_{1,\sigma}^{0°} + \sin(\theta) S_{1,\sigma}^{90°} \end{cases} \quad (6)$$

Where $\sigma$ represents the Gaussian standard deviation, and * denotes convolution operation.

Furthermore, in order to enhance the detailed information of images, thus the second-order gradients based on the three basic filters (i.e., $G_{2,\sigma}^{0°}$, $G_{2,\sigma}^{60°}$ and $G_{2,\sigma}^{120°}$) are applied in the construction process of the second-order channels. Similarly, the second-order steerable channels of the image $I(x, y)$ at an arbitrary orientation $\theta$ can be computed by convoluting the image with $G_{2,\sigma}^{0°}$, $G_{2,\sigma}^{60°}$ and $G_{2,\sigma}^{120°}$, which is expressed as Eq. (7).

$$\begin{cases} S_{2,\sigma}^{0°} = G_{2,\sigma}^{0°} * I(x, y) \\ S_{2,\sigma}^{60°} = G_{2,\sigma}^{60°} * I(x, y) \\ S_{2,\sigma}^{120°} = G_{2,\sigma}^{120°} * I(x, y) \\ S_{2,\sigma}^{\theta} = \cos^2(\theta)S_{2,\sigma}^{0°} + \sin^2(\theta)S_{2,\sigma}^{60°} - 2\sin(\theta)\cos(\theta)S_{2,\sigma}^{120°} \end{cases} \quad (7)$$

Once the synthetical first- and second-order steerable channels are constructed, the specified direction features at different scales are summed to obtain as much useful information as possible in each direction. Subsequently, these synthetical steerable channels in specified directions are convoluted by three parallel Dilated (or Atrous) Gaussian kernels, then the three parallel convolutional results are combined through one summation operation, which is designed to integrate a wealth of local inter-pixel information of images. The dilated Gaussian convolution with different dilated rates by inserting "holes" in the convolution kernels to expand its receptive field, which is inspired by the recent deep convolutional neural networks (Chen et al., 2017a; Chen et al., 2017b). In addition, the dilation rates $r$ are set to [ 1, 2, 3] for avoiding the inherent "gridding" problem that exists in the current dilated convolution framework (Wang et al., 2018a). By this means, the multiscale context structure features of the synthetical first- and second-order steerable channels can be further enhanced by utilizing dilated Gaussian weighting without increasing the computational complexity, and play a role in smoothing noise as well.

Fig. 3 clearly illustrates the advantages of utilizing the dilated Gaussian convolution for the construction of the SFOC descriptor. Four different types of heatmaps concerning different features are acquired by performing template matching. It is obvious that the heatmap of the original image pairs is the messiest, and the heatmap of the SFOC features without Gaussian convolution has several peaks but the peak is not distinct, because it's greatly affected by significant noise. In contrast, Gaussian convolution can effectively resist the interference of noise and make the peak more discriminative (see Fig. 3 (e) and (f)). Furthermore, the dilated Gaussian convolution can not only smooth the noise, but also further enhance the multiscale context structure features by the dilated Gaussian weighting. This is the reason why the heatmap of the SFOC features with parallel Dilated Gaussian convolution presents a smoother and more discriminative peak than the general Gaussian convolution, which indicates that the matching robustness of SFOC with parallel Dilated Gaussian convolution may be superior.

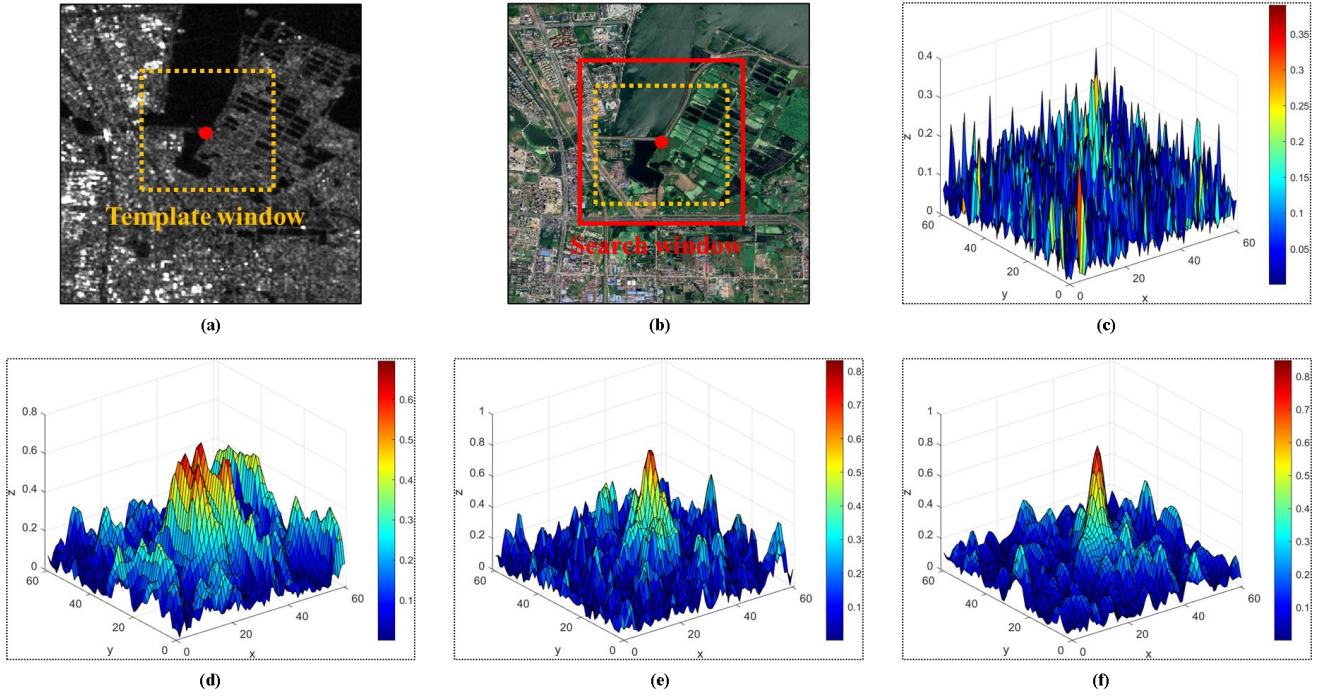

Fig. 3. Illustration of the constructed descriptor utilizing different Gaussian convolution strategies. (a) SAR image. (b) optical image. (c) Heatmap of the original image pairs. (d) Heatmap of the SFOC features without Gaussian convolution. (e) Heatmap of the SFOC features with the general Gaussian convolution. (f) Heatmap of the SFOC features with parallel Dilated Gaussian convolution.

In particular, for the synthetical second- order steerable channels, $\sigma$ with a larger value than the synthetical first-order steerable channels is used for dilated Gaussian smoothing as the second-order gradient describes more image detail but is accompanied by an increase in noises. Subsequently, the first- and second-order steerable channels are normalized respectively, then the final feature representation of SFOC is obtained by stacking them.

**2.2 Establishment of fast similarity measure**

The traditional normalized cross-correlation (NCC) is widely applied to determine corresponding CPs between the given image pairs with overlapping regions by evaluating the intensity similarity. However, it is often used only for CP detection of single-modal images, and is often unable to keep the same performance for multimodal image matching. As mentioned above, the SFOC descriptor can capture the structural features of images, which effectively resists NRD between multimodal images. Accordingly, it makes sense to establish a novel similarity measure that combines NCC with the SFOC descriptor.

SFOC is a 3D descriptor with a large amount of data, as well as the NCC also has the disadvantage of large calculation amount and high computational complexity. Hence, in order to maintain the matching accuracy and improve the computational efficiency, a fast-matching similarity measure is designed based on NCC and SFOC, it's expressed as Fast-NCC$_{SFOC}$. The proposed Fast-NCC$_{SFOC}$ can be reformulated with more detail as follows in this subsection.

First of all, the SFOC descriptor is used to calculate the structural features in the template image and the search image, which are denoted by $T$ and $S$, respectively. Their normalized correlation value $NCC_{SFOC}(S, T)$ represents the similarity of the template window $T(i, j, z)$ and the search window $S(x, y, z)$ at the location (x, y), which is defined as.

$$NCC_{SFOC}(S,T) = \frac{\sum_{h=1}^{z}\sum_{i=1}^{m}\sum_{j=1}^{n}[S(x+i,y+j,z)-\overline{S}][T(i,j,z)-\overline{T}]}{\sqrt{\sum_{h=1}^{z}\sum_{i=1}^{m}\sum_{j=1}^{n}[S(x+i,y+j,z)-\overline{S}]^2 \sum_{h=1}^{z}\sum_{i=1}^{m}\sum_{j=1}^{n}[T(i,j,z)-\overline{T}]^2}} \quad (8)$$

Where $z$ presents the dimension of the SFOC descriptor. $S(i, j, z)$ and $T(i, j, z)$ are the feature value of the search window and the template window at the position $(i, j, z)$, respectively. The sizes of the template and search window are $m \times n \times z$ pixels and $M \times N \times z$ pixels, respectively. $\overline{T}$ represents the average feature value of the template image, and $\overline{S}$ represents the average feature value of the search image $S$ under the current template image $T$.

The reason for the high computational complexity of traditional correlation matching is that NCC is completely recalculated for any search position $(x, y)$, while the internal relation of the NCC of adjacent search points is ignored. In order to reduce the computational complexity, an equivalent transformation is performed on Eq. (8), as follows:

$$NCC_{SFOC}(S,T) = \frac{R_{ST}(x,y,z) - R_S(x,y,z)R_T(i,j,z)/mnz}{\sqrt{[R_{SS}(x,y,z) - R_S^2(x,y,z)/mnz]}\sqrt{[R_{TT}(i,j,z) - R_T^2(i,j,z)/mnz]}} \quad (9)$$

There are only three items related to $(x, y, z)$ are included in the above formula, which is respectively denoted as:

$$R_{ST}(x,y,z) = \sum_{h=1}^{z}\sum_{i=1}^{m}\sum_{j=1}^{n} S(x+i,y+j,z)T(i,j,z) \quad (10)$$

$$R_S(x,y,z) = \sum_{h=1}^{z}\sum_{i=1}^{m}\sum_{j=1}^{n} S(x+i,y+j,z); R_{SS}(x,y,z) = \sum_{h=1}^{z}\sum_{i=1}^{m}\sum_{j=1}^{n} S^2(x+i,y+j,z) \quad (11)$$

$$R_T(i,j,z) = \sum_{h=1}^{z}\sum_{i=1}^{m}\sum_{j=1}^{n} T(i,j,z); R_{TT}(i,j,z) = \sum_{h=1}^{z}\sum_{i=1}^{m}\sum_{j=1}^{n} T^2(i,j,z) \quad (12)$$

It should be noticed that the first term $R_{ST}(x, y, z)$ in the numerator is convolution operation, and the convolution in the spatial domain is equivalent to the dot production operation in the frequency domain. Therefore, it can be converted to the frequency domain, and the FFT technique is used to improve computational efficiency. Accordingly, the new expression of the term is equivalent to the following form:

$$R_{ST}(x, y, z) = \int^{-1}[\int(S)\int^{*}(T)] \tag{13}$$

Where $\int$ is the signal of the Fourier transform, $\int^{*}$ represents the conjugate complex operation of the transformed result, and $\int^{-1}$ denotes the inverse FFT (i.e., IFFT).

Additionally, the terms in the denominator and the other terms in the numerator of Eq. (9) require a lot of multiplications and additions. When the template is sliding, the sum of the squares and correlation values are recalculated, which results in computation time increased enormously. It can be seen that these terms, $R_S$ and $R_{SS}$, fit the definition of the integral image (Viola and Jones, 2001). While the other two terms, $R_T$ and $R_{TT}$, are only related to the template image, which results in their values being fixed. Therefore, the integral image is used to replace the original summation process with three simple addition and subtraction operations, which can effectively reduce the computational complexity of the original algorithm to calculate NCC, and improve the running time.

The integral image $G(x, y, z)$ is an intermediate image representation whose feature values at the position $(x, y)$ is equal to the sum of the pixel values above and to the left of $(x, y)$, including the values at positions x, y, which can be represented as follows:

$$G(x, y, z) = \sum_{h=1}^{z}\sum_{i=1}^{x}\sum_{j=1}^{y} g(i, j, z) \tag{14}$$

Where $z$ presents the dimension of the SFOC descriptor, $g(i, j, z)$ is the SFOC feature, and $G(x, y, z)$ represents the corresponding integral image.

The integral feature map can be obtained in a single pass over the original feature by taking advantage of the following equations that are a pair of recursive formulas:

$$\begin{cases} s(x, y, z) = s(x, y-1, z) + g(x, y, z) \\ G(x, y, z) = G(x-1, y, z) + s(x, y, z) \end{cases} \tag{15}$$

where $s(x, y, z)$ represents the cumulative column sum. Since only two additions are required for each pixel, the computation of the integral image of $M \times N \times z$ pixels requires only $2MNz$ additions. A visualization of an integral image is shown in Fig. 4. Once the integral image is calculated, the integral of the feature values of the rectangular region with $(x, y)$ as the upper left vertex is:

$$\sum_{h=1}^{z}\sum_{i=x}^{m}\sum_{j=y}^{n} G(x,y,z) = G(x+m, y+n, z) - G(x, y+n, z) - G(x+m, y, z) + G(x, y, z) \tag{16}$$

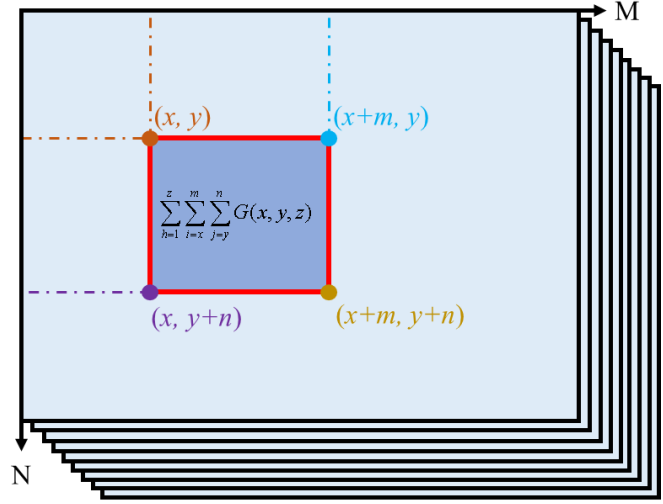

Fig. 4. Illustration of integral images. The black box represents the search window and the red box represents the template window

As a result, these terms, $R_S$ and $R_{SS}$, can be efficiently calculated utilizing the integral image by Eq. (16). Since the integral process only involves a limited number of additional operations, the complexity of the algorithm is mainly determined by FFT and IFFT in Eq.(13). Typical FFT and IFFT require about $2MNz\log_2(MNz)$ times of multiplication, and the Eq. (13) requires to calculate FFT and IFFT once in total. Accordingly, the total number of multiplications required by the proposed Fast-NCC$_{SFOC}$ is as follows.

$$T_1 \approx 4MNz\log_2(MNz) \tag{17}$$

With regard to the template matching strategy, the Eq. (8) is directly used to calculate NCC at each sliding position, and the calculation amount mainly depends on the dominant times of multiplication operation. For any search position, the Eq. (8) is used to calculate NCC for about three times of multiplication, and a total of $(M-m+1)\times(N-n+1)$ slidable positions need to be calculated for traversal search in the search window space. Thus, the number of multiplication operations required for NCC matching is:

$$T_2 = 3mnz(M-m+1)(N-n+1) \tag{18}$$

From the Eqs. (17) and (18), we can see that the computational complexity of the proposed Fast-NCC$_{SFOC}$ is independent of the template size, whereas the computational complexity of NCC is approximately proportional to the product of the template size and the search size, especially when $m$ and $n$ are small relative to $M$ and $N$. The ratio of the computational complexity of the two similarity measures is:

$$T = \frac{T_1}{T_2} \approx \frac{4MNz\log_2(MNz)}{3mnz(M-m+1)(N-n+1)} \tag{19}$$

To facilitate the illustration of the computational advantage of Fast-NCC$_{SFOC}$, we assume that $z=9$, $M=N$, $m=n$, and $M=2m$. The curve of $T$ changing with $m$ is shown in Fig. 5. As the template and search sizes increase, the ratio of the computational complexity between Fast-NCC$_{SFOC}$ and NCC decreases rapidly, that is, the larger the template and search sizes are, the greater the computational advantage of Fast-NCC$_{SFOC}$ is. Taking a template window $m=100$ as an example accompanied by a search window $M=200$, Fast-NCC$_{SFOC}$ takes about 0.965% of the time required by NCC, which greatly improves the computational efficiency.

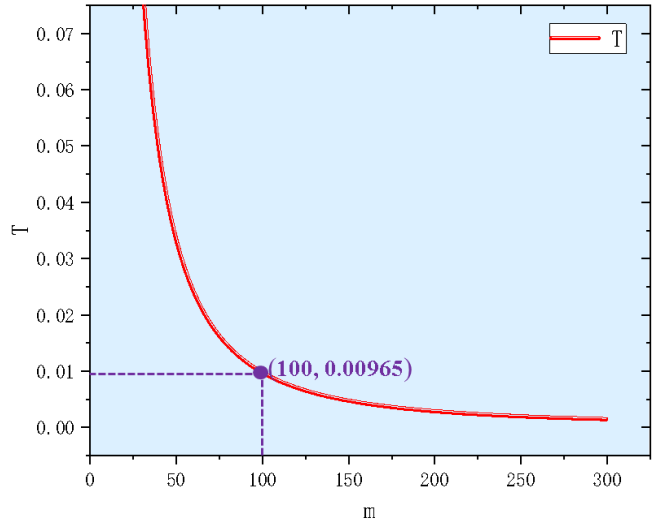

Fig. 5. Graph of the variation of T with m

## 3. REGISTRATION SYSTEM BASED ON THE PROPOSED MATCHING METHOD

In the previous section, SFOC is constructed to resist significant NRD between multimodal images, and Fast-NCC$_{SFOC}$ is designed for fast matching using FFT and integral images. These key steps are integrated into our multimodal image registration system. In addition to that, the registration system should further meet the following basic requirements: (1) A sufficient number of CPs need to be uniformly distributed because the number and distribution of CPs greatly directly affect the accuracy and quality of subsequent image registration. (2) An outlier rejection process is used to remove the mismatches caused by some occlusions (i.e., clouds and shadow).

The proposed registration system primarily carried out the following process to meet these requirements mentioned above: (1) local coarse registration, and (2) fine registration. Specifically, the local coarse registration consists of two steps: detection of interest points and local geometric correction. The fine registration involves three steps: (1) features extraction using SFOC, (2) CP detection using Fast-NCC$_{SFOC}$, and (3) outlier detection and image rectification. The workflow of the designed registration system is shown in Fig. 6, which is then described in detail.

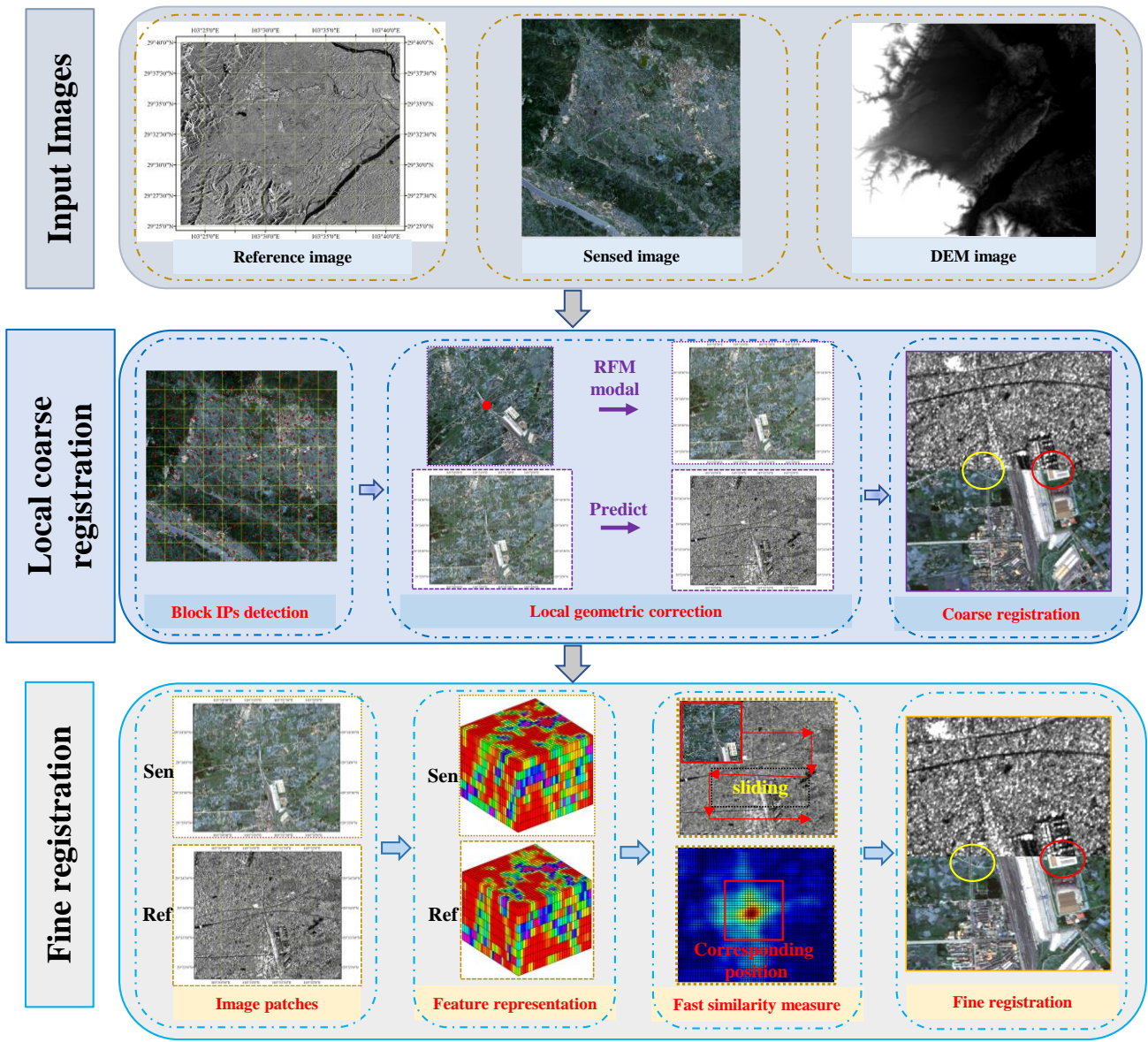

Fig. 6. The workflow of the proposed system

## 3.1 Local coarse registration

Most RS images, such as current state-of-the-art Earth observation data (e.g, TerraSAR, Sentinel, a series of the ZiYuan and GaoFen satellites, etc), usually have associated geo-referencing information, which can be applied to limit the search region to a smaller area for guiding the matching process. Accordingly, for different types of georeferencing images (i.e., L1 and L2 data), we designed two different strategies of coarse registration in the proposed system. One is the reference image has been geometrically corrected, while the sensed image comes with a file that includes the RPCs, where RFM is first employed to perform a local coarse correction to eliminate scale and rotation differences between the reference and sense images. The other is that both the reference and sensed images have been geometrically corrected, and the search area of image matching can be roughly predicted through geographic coordinates. In addition, if the resolution of the images is different, the higher resolution image should be uniformly sampled to the lower resolution image. The purpose of the coarse registration is to provide reliable spatial geometry constraints for the subsequent fine registration.

### 3.1.1 Detection of interest points (IPs)

The FAST operator is employed to detect IPs in our system because of its high computational efficiency, especially towards large-size RS images. However, the uneven distribution of the extracted IPs is a universal phenomenon (see Fig. 7a) when the FAST operator is directly applied on the whole image. It is well known that the uneven distribution of IPs could influence the fitting effect of the geometric transformation model, resulting in the registration accuracy degraded.

Accordingly, and given the importance of this issue, the designed system employs a partitioning strategy to extract the uniformly distributed interest points of the sensed image, which is named the block-based FAST operator (Ye et al., 2021b). As shown in Fig. 7b, compared with the original FAST operator (see Fig. 7a), the block-based FAST operator can detect a sufficient number of IPs with uniform distribution.

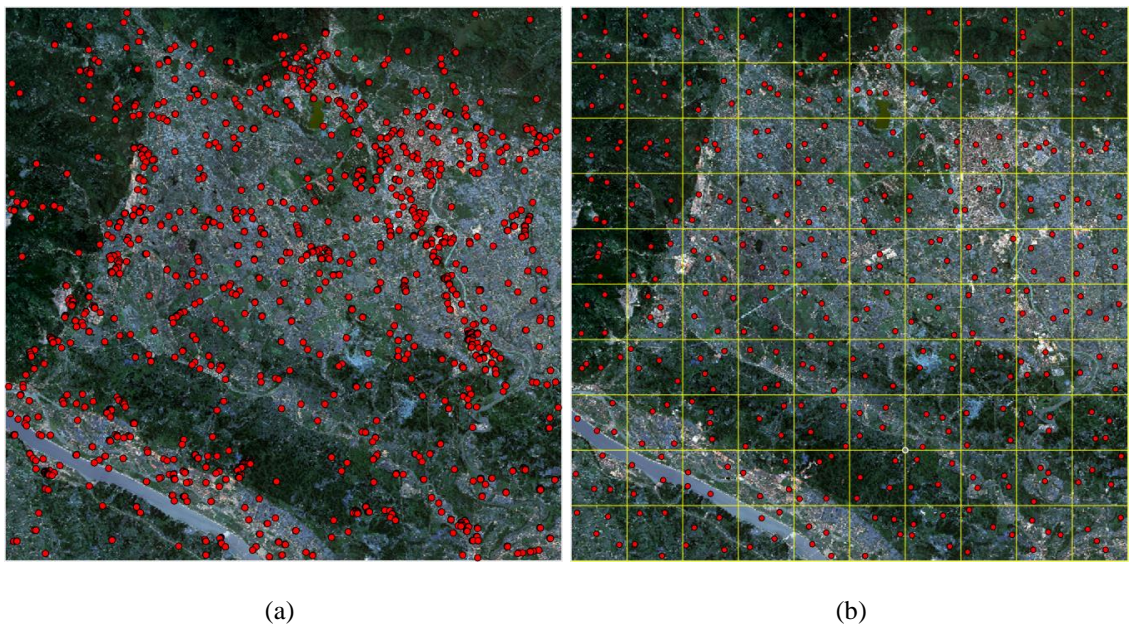

(a)                                             (b)

Fig. 7. Comparison of IP detection using different strategies. (a) IP detection by the original FAST operator. (b) IP detection by the block-based FAST operator.

## 3.1.2 Local geometric correction

After extracting the uniformly distributed IPs in the sensed image, an image patch (i.e., a template region) centered on an IP is first selected. If the sensed image comes with a file consisting of RPCs (i.e., L1 data), RFM is used to make the local geometric correction of the image patch, so that there is only a translation relationship between the patches (see Fig. 8).

The RFM model relates the normalized 3D ground coordinates (Latitude, Longitude, Height) to normalized 2D image pixel coordinates (line, sample) in the form of a ratio of cubic polynomials. In order to improve the stability of the polynomial equation coefficients, the 2D image coordinates and 3D ground coordinates are each offset and scaled to normalize within the range (–1.0, 1.0) (Wang et al., 2017). The general RFM model is defined as follows:

$$\begin{cases} r_n = \dfrac{Num_L(X_n, Y_n, Z_n)}{Den_L(X_n, Y_n, Z_n)} \\ c_n = \dfrac{Num_S(X_n, Y_n, Z_n)}{Den_S(X_n, Y_n, Z_n)} \end{cases} \quad (20)$$

Where $(r_n, c_n)$ are the normalized 2D image coordinates, and $(X_n, Y_n, Z_n)$ are the normalized 3D ground coordinates. $Num_L$, $Den_L$, $Num_S$ and $Den_S$ denotes the terms of third-order polynomials of $(X_n, Y_n, Z_n)$.

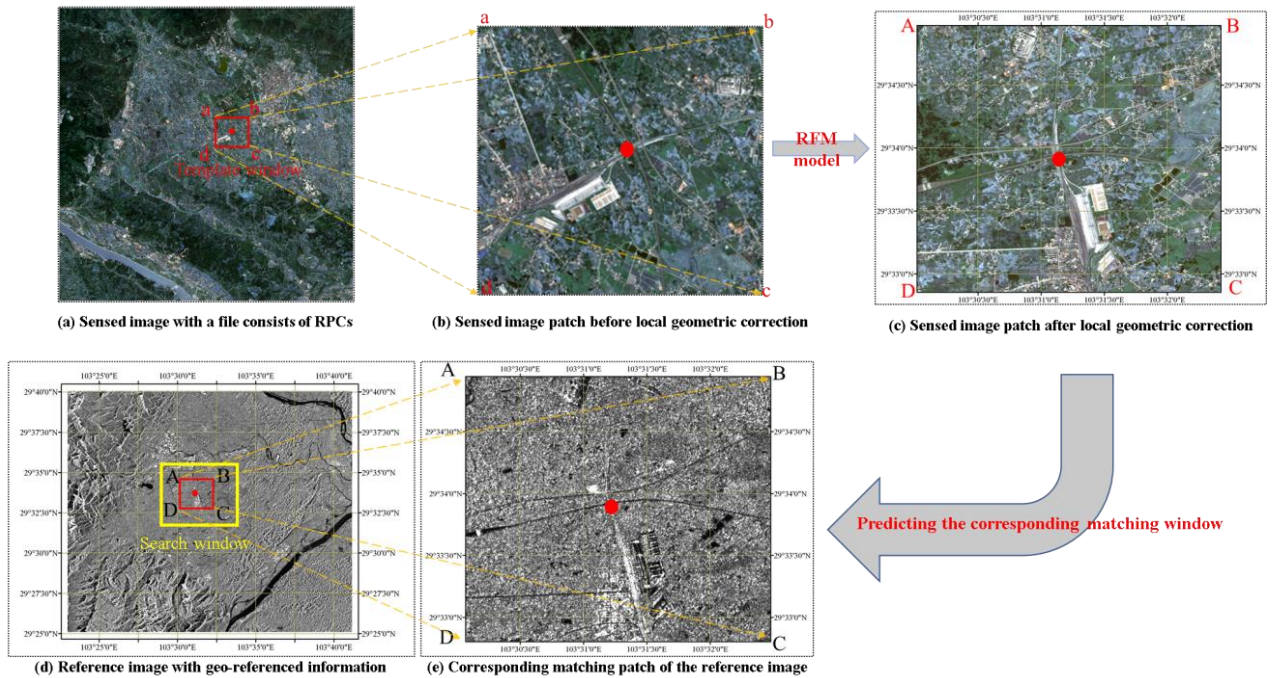

Fig. 8. Schematic of local geomatic correction

As shown in Fig. 8, the local geometric correction based on RFM can effectively eliminate obvious scale and rotation differences between the patches and reduce the matching area to a certain range. Moreover, the strategy of local coarse registration can effectively reduce the memory footprint of the designed system, as well as improve the matching efficiency, compared to performing the coarse correction for the entire sensed image.

In addition, if the sensed image has been geometrically corrected (i.e., L2 data), the corresponding matching range is directly predicted on the reference image based on the geographic coordinates, which is shown in Fig. 9. Furthermore, the higher resolution is resampled towards the lower resolution image if the resolution of the two images is different.

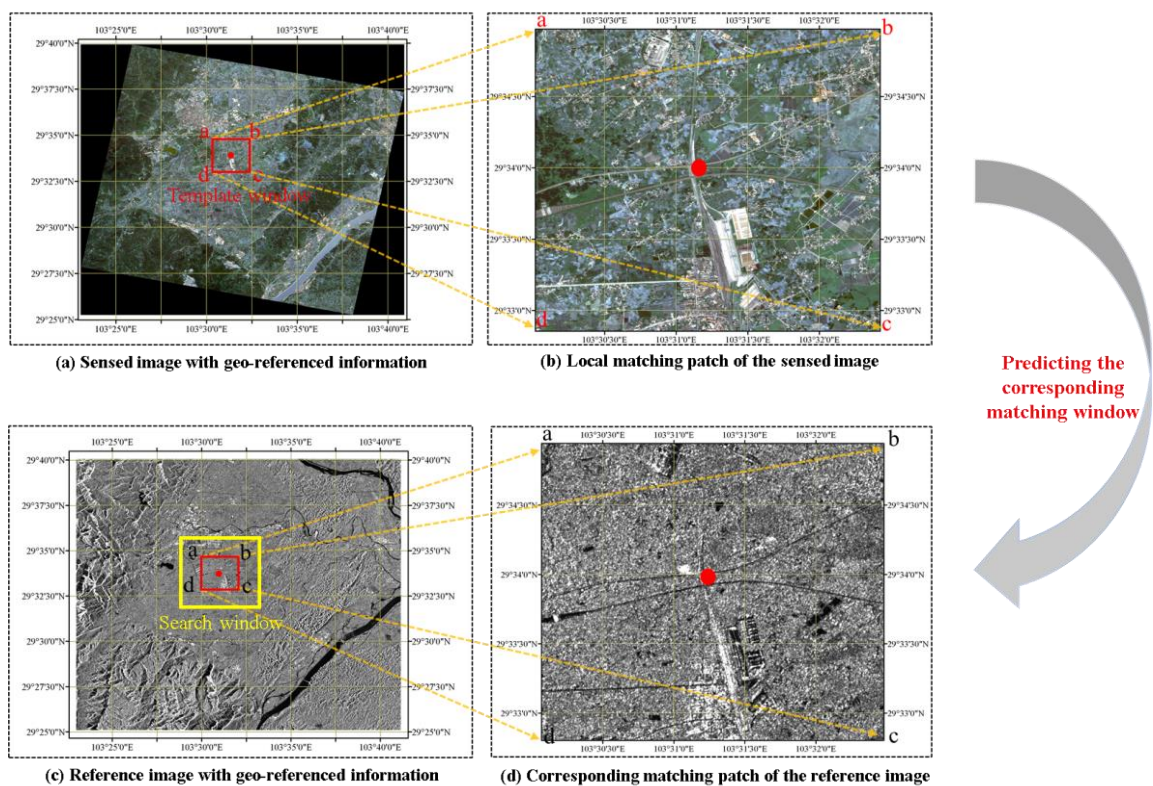

Fig. 9. Schematic of the sensed and reference image with geographical coordinates

### 3.2 Fine registration

By exploiting the geo-referenced information, it is reasonable to assume that the sensed and reference image are offset by only a few dozen pixels. After that, CPs are detected using Fast-NCC$_{SFOC}$ by a template matching scheme. Subsequently, RANSAC is employed to remove the CPs with large errors.

### 3.2.1 Features extraction using the SFOC descriptor

Since significant rotation and scale differences between image patches have been eliminated after the local geomatic correction, then the greatest matching difficulties are from severe NRD. Given that structure features have been widely used in multimodal image registration due to their robustness to NRD, our system employs the proposed SFOC descriptor for image matching. Fig. 10 visualizes the SFOC descriptors of a sensed image patch with the local geometric correction and its corresponding reference image patch. It can be clearly observed that the two SFOC descriptors look quite similar between the two multimodal image patches, which also intuitively illustrate that the subsequent steps are feasible to use Fast-NCC$_{SFOC}$ as the similarity measure.

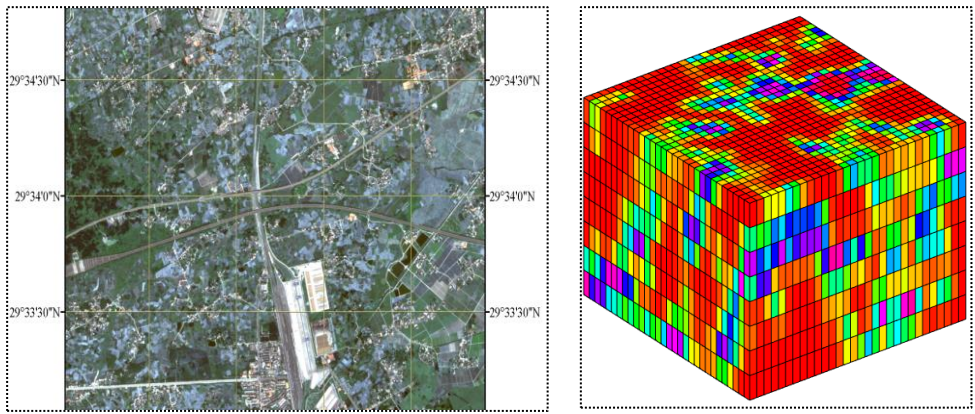

(a) Sensed image patch with the local geomatic correction and its SFOC feature representation.

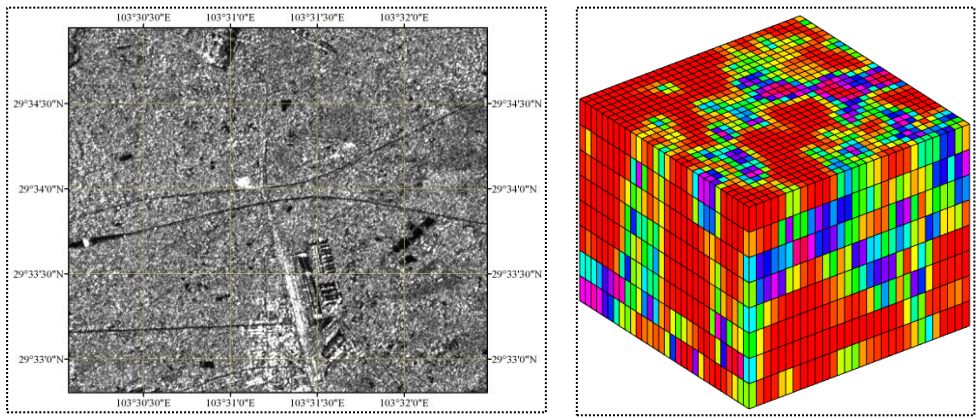

(b) Reference image patch and its SFOC feature representation.

Fig. 10. Visualization of the SFOC descriptor.

### 3.2.2 CP detection using Fast-NCC$_{SFOC}$

Given the similarity characteristics of the SFOC descriptors between multimodal images, the designed system uses Fast-NCC$_{SFOC}$ (see section 2.3) as the similarity measure for fast CP detection. The specific steps are as follows.

Firstly, a template window is selected around one IP from the sensed image, and the local geomatic correction is implemented for the template patch. Then the corresponding search window is predicted in the reference image on the basis of the geo-referenced information. Subsequently, SFOC is used to extract the structural features both in the template and the search window, respectively. Finally, Fast-NCC$_{SFOC}$ is employed for template matching, the position with the most similar SFOC feature is regarded as the CP (see Fig. 11).

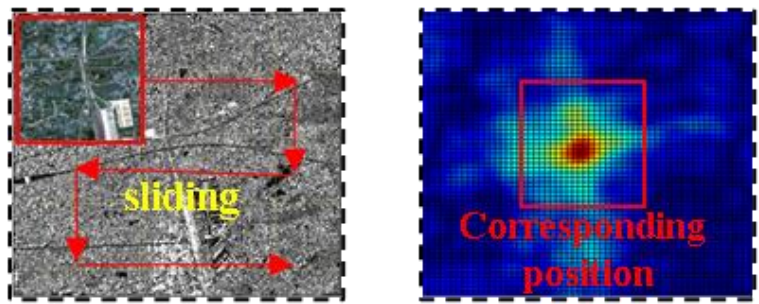

Fig. 11. Diagram of the template matching

### 3.2.3 Outlier detection and image rectification

Considering that some uncertain factors such as noise interference, distortions, and partial occlusion (e.g., clouds and shadow), mismatches are inevitable in practice. Thus, once the matching process is complete, RANSAC is used to remove the mismatches to obtain reliable CPs.

The advantages of RANSAC are its robust estimation of model parameters and its ability to estimate high-precision parameters from data sets containing a large number of mismatched CPs. In the designed system, when the sensed image comes with a file that includes RPCs, the image compensation scheme (Wang et al., 2017), combining the affine transformation and the RFM model defined by Eq. (21), is utilized to reject mismatches with an iterative strategy. For the other case, the estimation model is the projection transformation that is suitable for common geometric distortions with RS images.

$$\begin{cases} r + \Delta r = r + a_0 + a_1 r + a_2 c = RFM_r(X_N, Y_N, Z_N) \\ c + \Delta c = c + b_0 + b_1 r + b_2 c = RFM_c(X_N, Y_N, Z_N) \end{cases} \quad (21)$$

Where $(r,c)$ are the 2D image coordinates of CPs in the sensed image, $(\Delta r, \Delta c)$ represents the compensation values of systematic errors, $(a_1, a_2, a_3)$ and $(b_1, b_2, b_3)$ are coefficients of the affine compensation model, $(X_N, Y_N, Z_N)$ are the corresponding 3D ground coordinates of matched CPs in the reference image, and $RFM$ denotes the RFM model [see Eq. (20)].

After eliminating the mismatched CPs, the sensed image is corrected into the specified coordinate system of the reference image for image alignment. Considering both the accuracy and efficiency, the bilinear interpolation method is employed to resample the sensed image in the designed system.

## 4. EXPERIMENTAL EVALUATION: PERFORMANCE OF THE PROPOSED MATCHING METHOD

In this section, the performance of the proposed SFOC was experimentally evaluated with different types of multimodal RS datasets (e.g., optical, infrared, LiDAR, SAR, and rasterized maps). Firstly, the experimental settings were presented, which include the detailed information of all the test datasets, the evaluation criteria, the implementation details, and the parameters predefined. Then, SFOC was compared with the five state-of-the-art similarity measures (including MI, matching by tone mapping (MTM) (Hel-Or et al., 2013), PCSD, CFOG, and SDFG) for verifying its effectiveness. Finally, we analysed the robustness of SFOC against Gaussian white noise and speckle noise.

### 4.1 Experimental settings

Ten cases of multimodal image pairs with significant NRD were employed to evaluate the performance of SFOC. These cases cover a variety of low (10m), medium (2m-3m), and high (0.5-1.5m) resolutions, specifically consisting of two Optical-to-Infrared cases, three LiDAR-to-Optical cases, three Optical-to-SAR cases, and two Optical-to-Map cases. The detailed information of these cases is given in Table 1, and these image pairs of each case are displayed in Fig. 13. In addition, the two images of each case have been pre-registered with the same ground sample distance (GSD) to remove obvious rotation and scale differences.

Table 1. Detailed information of all test cases

| Category | Case | Image source | GSD | Data | Size | Location |
|---|---|---|---|---|---|---|
| **Optical -to- Infrared** | 1 | Daedalus optical | 0.5m | 04/2000 | 512×512 | Urban |
| | | Daedalus infrared | 0.5m | 04/2000 | 512×512 | |
| | 2 | QuickBird visible | 2.4m | 05/2006 | 1028×1137 | Suburban |
| | | QuickBird infrared | 2.4m | 05/2006 | 1028×1137 | |
| **LiDAR -to- Optical** | 3 | LiDAR intensity | 2m | 10/2010 | 600×600 | Urban |
| | | WorldView-2 optical | 2m | 10/2011 | 600×600 | |
| | 4 | LiDAR intensity | 2m | 10/2010 | 621×617 | Urban |
| | | WorldView-2 optical | 2m | 10/2011 | 621×621 | |
| | 5 | LiDAR depth | 2.5m | 10/2010 | 524×524 | Urban |
| | | WorldView 2 optical | 2.5m | 10/2011 | 524×524 | |
| **Optical -to- SAR** | 6 | Sentinel-2 optical | 10m | 09/2018 | 1501×1501 | Suburban |
| | | Sentinel-1 SAR | 10m | 10/2018 | 1501×1501 | |
| | 7 | Google Earth | 3m | 11/2007 | 528×524 | Urban |
| | | TerraSAR-X | 3m | 12/2007 | 534×524 | |
| | 8 | Google Earth | 3m | 03/2009 | 628×618 | Suburban |
| | | TerraSAR-X | 3m | 01/2008 | 628×618 | |
| **Optical -to- Map** | 9 | Image from Google Maps | 0.5m | unknow | 700×700 | Urban |
| | | Map from Google Maps | 0.5m | unknow | 700×700 | |
| | 10 | Image from Google Maps | 1.5m | unknow | 621×614 | Urban |
| | | Map from Google Maps | 1.5m | unknow | 621×614 | |

In the experiments, the block-based FAST operator was first employed to extract 200 uniformly distributed IPs from the reference image. Then, the CP detection was performed using different similarity measures with a template matching manner. Considering that the larger the template window was, the better the matching performance was (Ye et al., 2019; Liang et al., 2021), the template window with the size of 80 × 80 pixels was used for all the similarity measures. Furthermore, four criteria were used to quantitatively evaluate the matching performance in terms of the number of correct match (NCM), the correct matching ratio (CMR), the root-mean-square errors (RMSE), and the matching time (MT). The correct match was determined by manually selecting 50 evenly distributed CPs to estimate the projective model for the image pairs of each case. The projective model was used to calculate the location errors of the matches obtained by different similarity measures, and the match within positioning errors of 1.5 pixels was defined as the correct CP. CMR was defined as CMR = NCM / total matches, where total matches refer to all matched CPs, including the outliers with large errors. The RMSE is expressed as:

$$RMSE = \sqrt{\frac{\sum_{i=1}^{N}\left(x_i^r - P(x_i^s, y_i^s)\right)^2 + \left(y_i^r - P(x_i^s, y_i^s)\right)^2}{N}} \tag{22}$$

Where $x_i^r$, $y_i^r$ and $x_i^s$, $y_i^s$ are the pixel coordinates of the correct CP $i$ between the reference and sensed image, $P$ represents the projective model and $N$ is the number of CPs.

Table 2. Parameters setting of all similarity measures

| Method | Parameters setting |
|--------|-------------------|
| **MI** | 32 histogram bins |
| **MTM** | Default settings |
| **PCSD** | Order partitions number: 3, angle interval number: 9, radius interval number: 4 |
| **CFOG** | Gaussian STD: 0.8, orientated gradient channel number: 9 |
| **SDFG** | Default settings |
| **SFOC** | the Gaussian STD: $\sigma_1 = 0.6$, $\sigma_2 = 0.8$, $\sigma_3 = 1$ and $\sigma_4 = 1.5$ |

To make a fair comparison, MI was calculated using a histogram with 32 bins, as this is usually accompanied by an optimal matching performance (Ye et al., 2019). And the parameters of the other comparative similarity measures (i.e., MTM, PCSD, CFOG, and SDFG) used the best parameters recommended in their related papers, which are given in Table 2. As aforementioned in Section 2.2, the performance of SFOC was related to two key parameters, i.e., the Gaussian STD ($\sigma_1, \sigma_2$, and $\sigma_3$) of first-order steerable channels, and the Gaussian STD ($\sigma_4$) of second-order steerable channels. Their influences had been tested by the multimodal image cases described in Table 1, which manifests SFOC with the parameters ($\sigma_1 = 0.6$, $\sigma_2 = 0.8$, $\sigma_3 = 1$ and $\sigma_4 = 1.5$) achieved the optimal matching capacity. All experiments were performed using a personal computer (PC) with the configuration of Inter (R) Core (TM) CPU i7-10750H 2.6GHz and 16GB RAM.

**4.2 Comparison and analysis of matching performance**

In this section, the performance of the proposed method was quantitatively and qualitatively evaluated. With the same IPs extracted by the block-based FAST detector, the quantitative evaluation was first performed by comparing SFOC with five state-of-the-art descriptors: MI, MTM, PCSD, CFOG, and SDFG. Moreover, in order to evaluate the effectiveness of the second-order gradient in the generation process of SFOC, the SFOC descriptor was degraded by only using the first-order steerable channels without the second-order steerable channels. The degraded SFOC descriptor was represented by F-SFOC, and it was also used for matching performance comparison with other similarity measures.

The seven different similarity measures, i.e., MI, MTM, PCSD, CFOG, SDFG, F-SFOC, and SFOC, were applied to ten multimodal image cases (Table 1) for the comparison of matching performance. Fig. 12 depicts the comparison results of all the evaluation criteria (i.e., NCM, CMR, RMSE, and MT) for the different methods on each multimodal image pair. It is obvious that SFOC outperformed the other methods for the above four criteria in all test cases, which effectively demonstrates the superiority and robustness of the proposed SFOC.

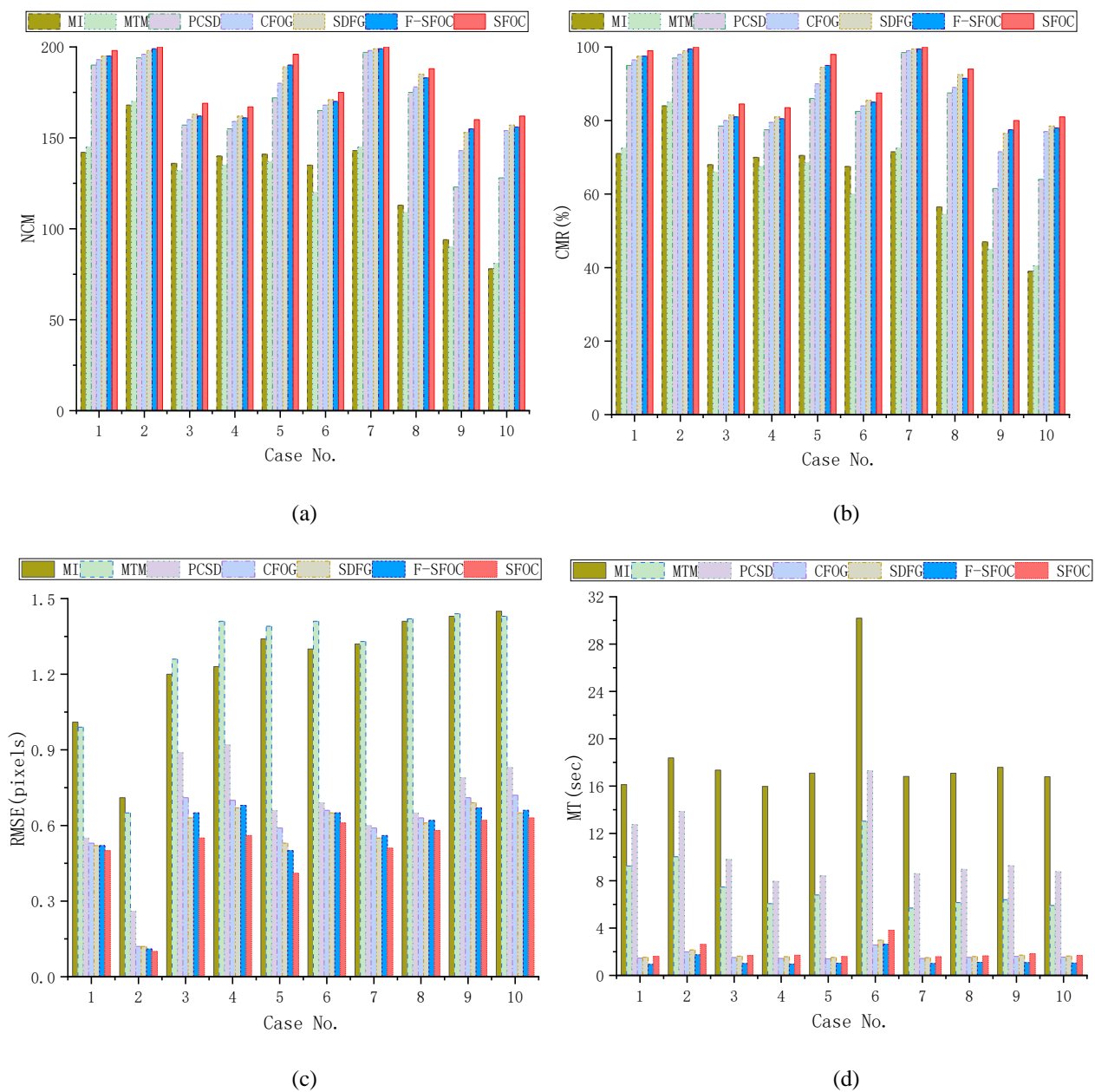

Fig. 12. Performance comparison of different methods on the ten multimodal image cases with the template size of 80 × 80 pixels. (a) NCM. (b) CMR. (c) RMSE. (d) MT.

Among the six similarity measures used for comparison, the worst matching performance was found in the MI and MTM. MI and MTM had comparable matching performance, but MTM performed slightly better than MI on two Optical-to-Infrared cases, while MI performed better than MTM on cases 3-8. This may be related to the fact that MTM only utilizes a piecewise linear function to fit the intensity changes widely existing in the multimodal images. However, the intensity relationship between optical and SAR (or LiDAR) images is too complex to be fitted by MTM, which results in its performance degradation. Although the performance of MI was slightly better than that of MTM, it was the most time-consuming among all the similarity measures because it requires calculating a large number of joint probability histograms.

From the comparison results in Fig. 12, we can also observe that PCSD, CFOG, and SDFG performed significantly better than MI and MTM, while SDFG had slightly better performance compared with CFOG and PCSD. The main reason is that PCSD is constructed by using the multiscale phase congruency structural features, and CFOG is built making use of the dense channel features of orientated gradients, which is more robust to NRD than MI and MTM. When comparing PSCD with CFOG, its performance was slightly worse than that of CFOG. The reason for that is the PCSD may lose some detailed structural information because it employed the strategy of the phase congruency order-based region division for descriptor construction, As for SDFG, since it further increasingly adopted the multi-scale strategy on the basis of multi-direction using odd Gabor functions, its matching performance was more robust than CFOG, but the matching process was more time-consuming. In addition, the construction of PCSD relies on multiscale phase congruency features, which results in it being time-consuming. Therefore, PCSD and MTM were the most time-consuming apart from MI in all the compared similarity measures.

For our degraded descriptor (i.e., F-SFOC), its matching performance was comparable to SDFG, and it yielded better results than CFOG on the criterion of RMSE, especially in the LiDAR-to-Optical and Optical-to-SAR cases. This phenomenon illustrates that the first-order Gaussian steerable filters and the dilated Gaussian convolution are effective to construct the descriptor. While the matching performance of F-SFOC was obviously lower than SFOC, which verified the feasibility and effectiveness of adding the second-order gradient in the generation of SFOC. In this way, the robustness and discriminability of SFOC can be effectively increased. As far as the MT, F-SFOC was slightly faster than CFOG, because it only took advantage of the first-order steerable channels without the second-order steerable channels resulting in a smaller dimensionality of its features than that of CFOG. Whereas SFOC required slightly more time-consuming than CFOG and SDFG, this is related to the multi-scale strategy with different Gaussian STD and the dilated Gaussian convolution with different dilated rates were embedded in the generation process of SFOC. Hence, considering the improvement of the matching performance for SFOC, it is acceptable to sacrifice a little running time.

Moreover, qualitative evaluation was performed by displaying the correct matched CPs for the visual inspection. As shown in Fig. 13, these CPs were established by SFOC on the image pairs of each case with a template size of $80 \times 80$ pixels. it is obvious that these obtained CPs on the image pairs of each case were evenly distributed, and the location accuracy of these CPs was relatively precise despite significant NRD and noise between these multimodal image pairs.

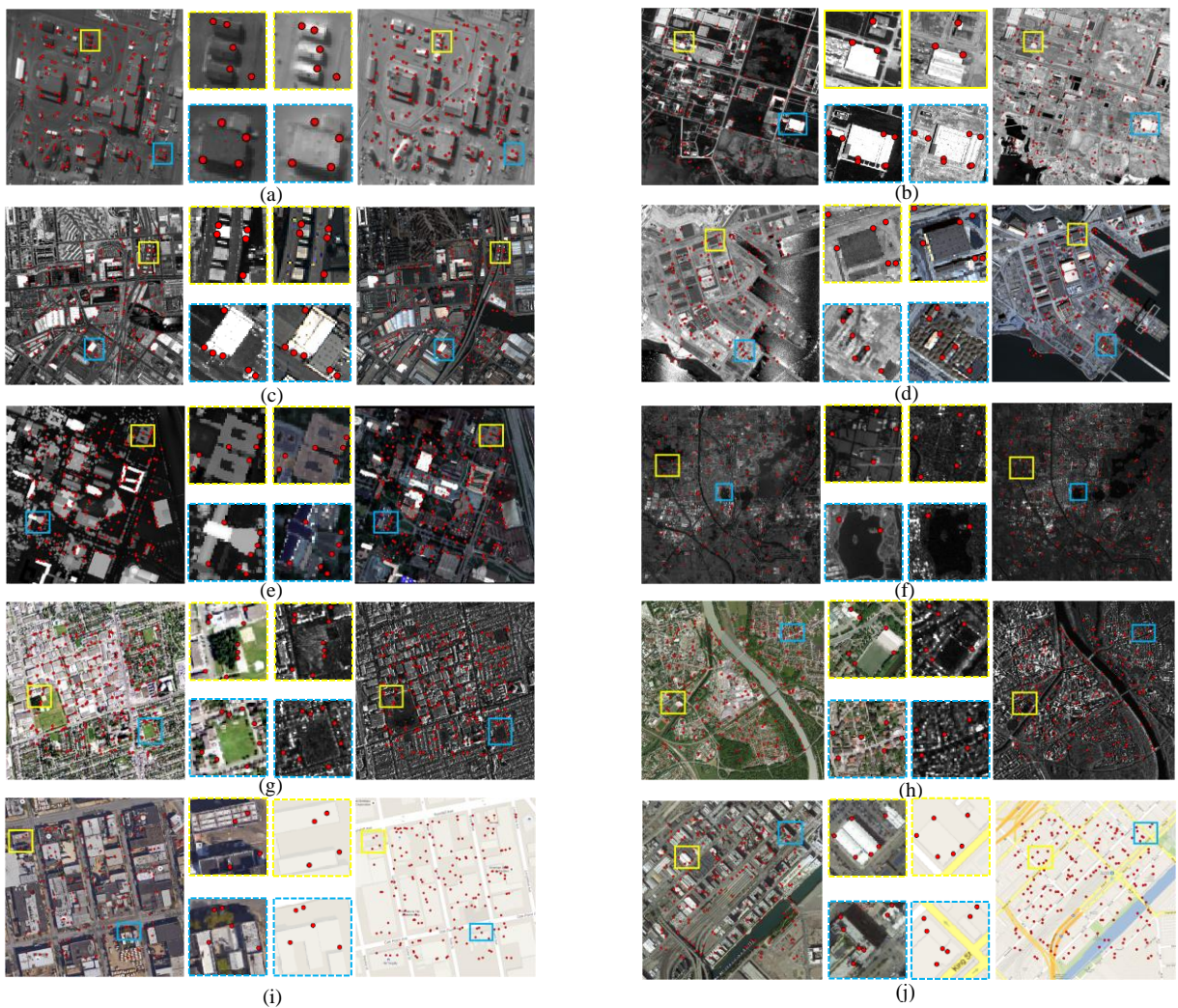

Fig. 13. Matching results of all test cases by SFOC with the template size of $80 \times 80$ pixels. (a) Case 1. (b) Case 2. (c) Case 3. (d) Case 4. (e) Case 5. (f) Case 6. (g) Case 7. (h) Case 8. (i) Case 9. (j) Case 10.

## 4.3 Comparison and analysis of noise sensitivity

In this section, the anti-noise performance of the above-mentioned similarity measures was evaluated and analyzed by adding different levels of Gaussian white noise and speckle noise to the images, respectively. Because the NRD between multimodal images is difficult to be precisely fitted only by a simple mathematical model. Meanwhile, LiDAR and SAR images typically contain more noise than infrared images, which is not conducive to the assessment of noise sensitivity. Consequently, all the similarity measures were performed with the template size of $80 \times 80$ pixels for the selected four pairs of Optical-to-Infrared cases, and their average value of CMR was used for the subsequent analysis. Specifically, two types of series noisy images were generated by adding the different levels of Gaussian white noise with mean 0 and variance $v$ in the range [0, 0.01] with an interval of 0.001, and the different levels of speckle-noise with variances $v$ in the range [0, 0.1] with an interval of 0.01, respectively.

Fig. 14 presents the average CMRs of different similarity measures versus various noise consisting of Gaussian white noise and speckle noise. SFOC and its degraded version (i.e., F-SFOC) achieved superior capacities under increasing Gaussian and speckle noise, followed by SDFG and CFOG. It demonstrated that the generation of SFOC using the dilated Gaussian convolution with different dilated rates could be more effective for resisting noise than SDFG only utilizing the general Gaussian convolution, and the generation of SFOC and SDFG both using a series of filters was more useful in withstanding noise than CFOG only utilizing simple gradient computation with the pixel difference. While the orientation channels of CFOG were implemented by the Gaussian kernel, which is more effective to reduce the interference of noise than PCSD. In addition, the performance of MI was relatively stable under various noises, but its average CMR was still lower than SFOC, F-SFOC, and CFOG. And MTM also presented lower robustness to Gaussian and speckle noise compared with MI.

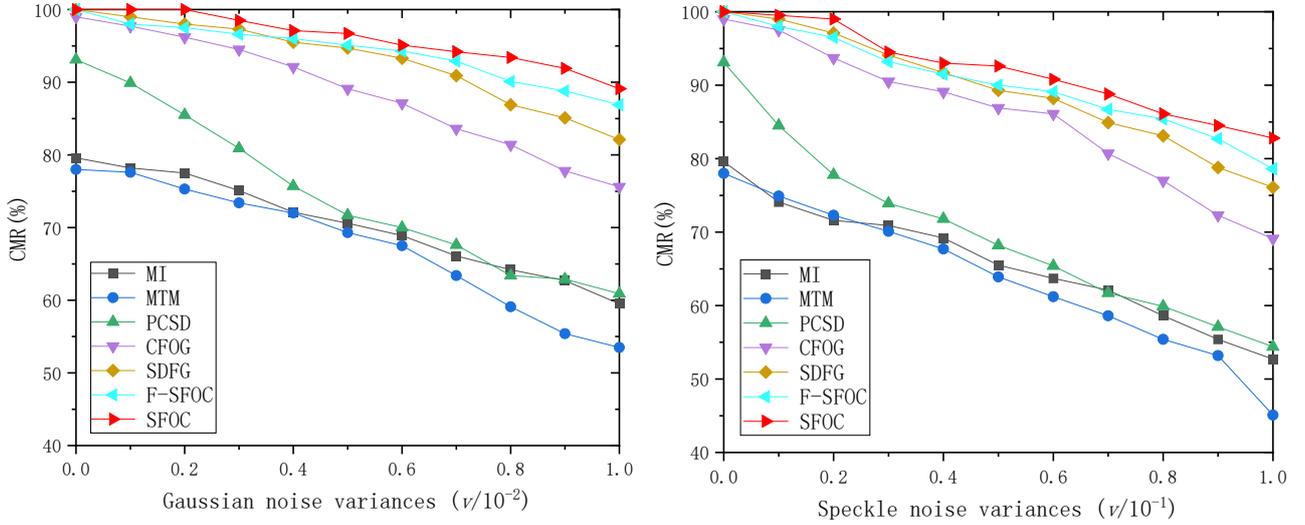

Fig. 14. Average CMRs of different similarity measures versus various noise. (a) Average CMRs of different similarity measures versus various Gaussian white noise. (b) Average CMRs of different similarity measures versus various speckle noise.

The above results and coherence analysis demonstrate that SFOC has apparent effectiveness and advantages for resisting significant NRD and noise between multimodal images, as well as high computational efficiency. The good adaptive performance was mainly due to the following reasons. On the one hand, it not only employed the first-order steerable filters with the multi-scale strategy to depict the multi-direction and multiscale structure features between multimodal images, but it also utilized the second-order steerable filters and three parallel dilated Gaussian kernels to emphasize more detailed structures, which further improves the discriminative and anti-noise capability of the proposed method. On the other hand, the improved Fast-NCC$_{SFOC}$ based on the FFT and integral image technique ensured its fast computational efficiency.

## 5. EXPERIMENTAL EVALUATION: PERFORMANCE OF THE DESIGNED REGISTRATION SYSTEM

In this section, the registration performance of the designed system was analyzed both in qualitative and quantitative evaluation, compared with the three popular commercial software systems (i.e., ENVI 5.3, ERDAS 2015, and PCI Geomatic 2016) by testing the same multimodal images. In the designed system, the programming language was C++ and the interface was designed by Qt. For image reading and writing, the implementation of the open-source Geospatial Data Abstraction Library (GDAL) in C++ was used. As is known to all, ENVI 5.3, ERDAS 2015, and PCI Geomatic2016 all have the automatic registration function modules, which have been widely used for remote sensing image registration. These function modules are called "Image Registration Workflow (ENVI)", "AutoSync Workstation (ERDAS)", and "OrthEngine-Automatic GCP Collection (PCI)", respectively. All the experiments were performed on the personal computer (PC) with the same configuration as described in the previous section (i.e., CPU i7-10750H 2.6GHz and 16GB RAM).

**5.1 Description of experimental datasets**

Experiments were conducted by applying the above three systems and the designed system to various multimodal images acquired at different geographic districts and times. The selected experimental datasets intentionally exhibit large image sizes, severe geometric distortions, significant NRD, and large differences in acquisition date and spatial resolution, and they make sense to testify the universality and robustness of the designed system.

Two different categories of testing datasets were used to evaluate our system. The first category is both the sensed and the reference images that had been geometrically corrected (i.e., L2 data), the second category is the sensed image with RPCs (i.e., L1 data). The summary descriptions of the experimental datasets are given in Table 3 and Fig. 15. Among them, test 1 and test 2 belong to the first category, and the second category consists of test 3 and test 4.

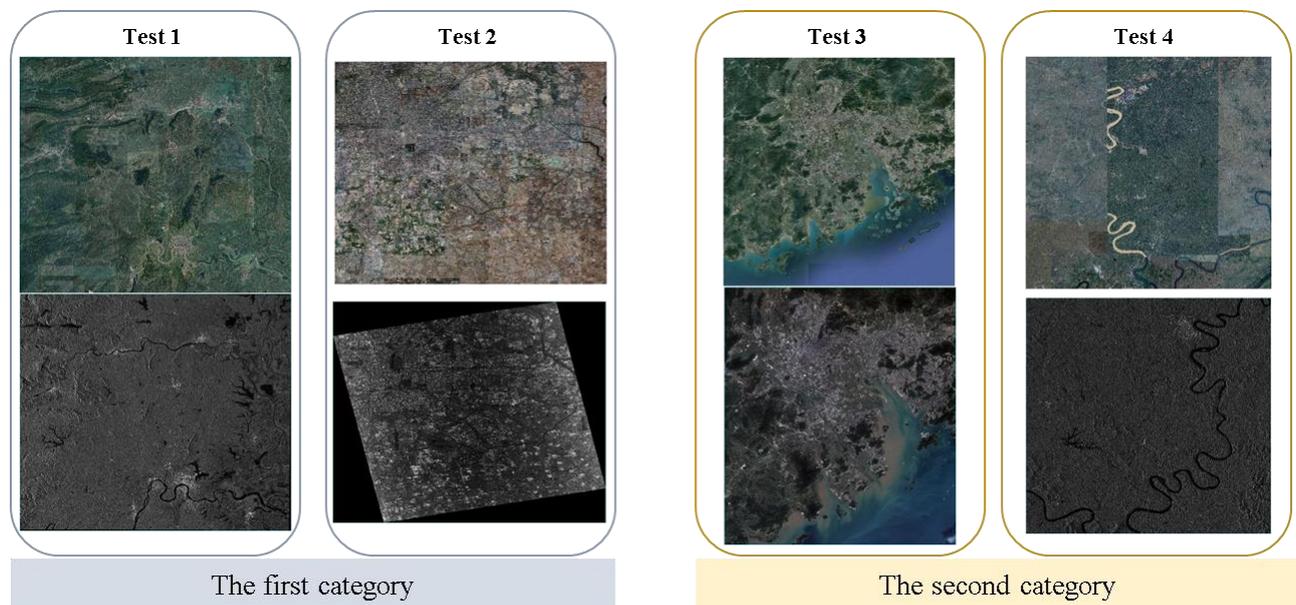

Fig. 15. The experimental test image pairs. In each pair of columns, the top is the reference image and the bottom is the sensed image.

Table 3. Summary descriptions of the test images

| Category | Test Case | Reference Image | Sensed Image | Image Characteristic |
|---|---|---|---|---|
| The First Category (L2 data) | Test 1 | Sensor: Google Earth, Optical<br>Spatial resolution: 4.5m<br>Data: 8/2016<br>Size: 26880×23552 | Sensor: Sentinel-1, SAR<br>Spatial resolution: 10m<br>Data: 09/2018<br>Size: 10980×10980 | The west of images is covered by mountains, while the east covers a suburban area with rivers and buildings. Moreover, the image resolution difference is about two times. |
| | Test 2 | Sensor: Google Earth, Optical<br>Spatial resolution: 2m<br>Data: 02/2018<br>Size: 22784×11688 | Sensor: GaoFen-3, SAR<br>Spatial resolution: 1.5m<br>Data: 07/2018<br>Size: 26442×21598 | Images cover an urban area with high buildings and have local geometric distortions. Moreover, there is obvious noise on the SAR image. |
| The Second Category (L1 data) | Test 3 | Sensor: Google Earth, Optical<br>Spatial resolution: 9m<br>Data: 01/2014<br>Size: 26624×26368 | Sensor: GaoFen-1, WFV<br>Spatial resolution: 16m<br>Data: 11/2019<br>Size: 12000×13400 | Images cover buildings, rivers, mountains, and seas. Moreover, there is a temporal difference of about five years in the local region, and the resolution difference is about two times. |
| | Test 4 | Sensor: Google Earth, Optical<br>Spatial resolution: 1m<br>Data: 04/2016<br>Size: 37376×35584 | Sensor: GaoFen-3, SAR<br>Spatial resolution: 3m<br>Data: 09/2020<br>Size: 21852×21241 | Images cover an undulating terrain with hills, rivers, and buildings. Moreover, there is obvious noise on the SAR image, and the resolution difference is about three times. |

## 5.2 Implementation details and evaluation criteria

As far as we are aware, ENVI, ERDAS, and PCI all adopt the template matching scheme to detect CPs between the sensed and reference images, which is similar to our system. In general, the best fitting template (i.e., the location of the CP) is identified by the similarity measure in the matching. For the selection of similarity measures, different commercial software employs different similarity measures. ENVI provided NCC and MI to carry out image matching, which was redefined as "$ENVI_{NCC}$" and "$ENVI_{MI}$" in the following sections, respectively. While PCI utilized NCC and Fast Fourier Transform Phase Matching (FFTP) to perform the matching process, which was referred to as "$PCI_{NCC}$" and "$PCI_{FFTP}$" in subsequent sections, respectively. Whereas ERDAS achieved CPs by NCC, as well as employed pyramid-based matching techniques to improve the matching performance, which was indicated as "$ERDAS_{NCC}$".

Furthermore, in the template matching process, the larger the window is, the higher the matching performance is, but also more time-consuming. Accordingly, the matching parameter setting of the software will affect the efficiency and precision of image matching. In order to make a fair comparison, the matching parameters of all the systems should be set the same as far as possible, and the image rectification was performed using the second-order polynomial model. Table 4 shows the common matching parameters between each system, these parameters were set to the same values in the comparison experiment.

Table 4. Common matching parameters in all the systems

| Parameter item | Designed system | ENVI | ERDAS | PCI |
|---|---|---|---|---|
| Similarity measure | Fast-NCC$_{SFOC}$ | NCC / MI | NCC | NCC / FFTP |
| Number of detected IPs | 400 | 400 | 400 | 400 |
| Search window size | 200 | 200 | Default | 300 |
| Template window size | 100 | 100 | Default | Default |

Whereafter, the experimental results were analyzed in terms of matching precision and computational efficiency. The matching precision was related to the NCM, CMR, and RMSE, and their definition was consistent with the previous section, which can be served as the evaluation criterion of quantified registration accuracy. Meanwhile, the matching time (MT) of the matching process was recorded to evaluate the running efficiency of different systems. In addition, we also provided the visualization of the matching CP pairs of each case and the superposition verification of the reference and sensed images, whereby the registration performance of the designed system can be verified intuitively and qualitatively.

**5.3 Comparison and analysis of registration performance**

Table 5 shows the quantitative comparison between the designed system and the other systems among the four criteria in terms of NCM, CMR, MT, and RMSE for the experimental image pairs of the first category. As far as the overall performance in Table 5 is concerned, the designed system achieved the best matching performance for all the image pairs of each case, which demonstrated the effectiveness of the designed system. Since diverse image pairs of each case had differences in geometric distortions, NRD, noise, and spatial resolution, each software system presented the different matching results of different cases.

The performance of the NCC-based (i.e., ENVI$_{NCC}$ and PCI$_{NCC}$) systems was the worst, with no CPs detected by these systems. Since ERDAS$_{NCC}$ employed the pyramid-based matching technique to enhance the robustness of matching, its overall performance was slightly better than ENVI$_{NCC}$ and PCI$_{NCC}$. In addition, there were a few CPs identified by PCI$_{FFTP}$ only for test 1, but their accuracy was poor. To sum up, the designed system presented the smallest RMSE, the fastest MT (except PCI$_{FFTP}$), the highest CMR, and the most NCM in all test cases, and the matching accuracy of ENVI$_{MI}$ was second best. The overall performance of ERDAS$_{NCC}$ ranked third, PCI$_{FFTP}$ fourth, and NCC-based last.

Table 5. Quantitative comparison for matching results of the first category

| Test Case | System | NCM | CMR (%) | MT (sec.) | RMSE (pixel) |
|---|---|---|---|---|---|
| Test 1 | ENVI$_{NCC}$ | Failed | Failed | Failed | Failed |
| | ENVI$_{MI}$ | 30 | 7.50% | 245.04 | 2.94 |
| | ERDAS$_{NCC}$ | 25 | 5.25% | 56.89 | 6.21 |
| | PCI$_{NCC}$ | Failed | Failed | Failed | Failed |
| | PCI$_{FFTP}$ | 13 | 3.25% | 12.61 | 9.63 |
| | Designed System | 307 | 76.75% | 23.45 | 1.21 |
| Test 2 | ENVI$_{NCC}$ | Failed | Failed | Failed | Failed |
| | ENVI$_{MI}$ | 20 | 5.00% | 435.29 | 3.53 |
| | ERDAS$_{NCC}$ | 19 | 4.75% | 53.32 | 6.76 |
| | PCI$_{NCC}$ | Failed | Failed | Failed | Failed |
| | PCI$_{FFTP}$ | Failed | Failed | Failed | Failed |
| | Designed System | 268 | 67.00% | 27.81 | 1.86 |

Table 6 shows the quantitative comparison between the designed system and other systems among the four criteria in terms of NCM, CMR, MT, and RMSE for the experimental image pairs of the second category. Since ERDAS does not have the automatic registration module for satellite images with RPCs, the "×" indicates that there was no registration function for such category in Table 6. The ENVI system only employed the NCC as a similarity measure to match CPs for the second type of images with RPCs, which was referred to as ENVI$_{NCC}$ that was compared with our system and PCI.

It was apparent from Table 6 that our system still achieved the best matching performance in the two tested cases of the second scheme. For test 3 where the matching performance of ENVI$_{NCC}$ and PCI$_{NCC}$ was still low, and the CMR was only 10.5% and 12.25%, respectively. And the matching performance of PCI$_{FFTP}$ is much better than PCI$_{NCC}$ in test 3. Whereas the CMR of our system was as high as 84.75% and the accuracy was much higher than ENVI$_{NCC}$, PCI$_{NCC}$, and PCI$_{FFTP}$. Moreover, ENVI$_{NCC}$, PCI$_{NCC}$, and PCI$_{FFTP}$ matching comprehensively failed for the Optical-to-SAR case of test 4, but the designed system still maintained a high CMR of 64.5%. These results quantitatively demonstrated that the local coarse-to-fine registration processes devised by our system were effective for matching multimodal images with RPCs. The local coarse registration with spatial geometric constraints using the RFM model can effectively eliminate the obvious geometric distortions between images. Moreover, the local geometric correction can avoid correcting the entire sensed image with RPCs that will yield an intermediate image file, as well as reduce the memory consumption of the system while ensuring the timeliness of our system.

Table 6 Quantitative comparison for matching results of the second category

| Test Case | System | NCM | CMR (%) | MT (sec.) | RMSE (pixel) |
|---|---|---|---|---|---|
| Test 3 | $ENVI_{NCC}$ | 42 | 10.50% | 25.88 | 2.75 |
| | ERDAS | × | × | × | × |
| | $PCI_{NCC}$ | 49 | 12.25% | 12.78 | 2.85 |
| | $PCI_{FFTP}$ | 299 | 74.75% | 17.62 | 1.25 |
| | Designed System | 339 | 84.75% | 23.46 | 0.87 |
| Test 4 | $ENVI_{NCC}$ | Failed | Failed | Failed | Failed |
| | ERDAS | × | × | × | × |
| | $PCI_{NCC}$ | Failed | Failed | Failed | Failed |
| | $PCI_{FFTP}$ | Failed | Failed | Failed | Failed |
| | Designed System | 258 | 64.50% | 23.06 | 1.69 |

The above different registration performance of these systems can be attributed to the following reasons. Firstly, the NCC and phase correlation methods were more sensitive to significant NRD between multimodal images. And MI had stronger NRD-suppressing abilities than the NCC and phase correlation methods, which ascribed its generation process using the entropy and joint entropy in information theory. Whereas, the designed system made use of SFOC to extract structural features, which enhanced the robustness of the designed system by depicting the multi-scale and multi-directional structural information. Meanwhile, the computational efficiency of the designed system was guaranteed by using Fast-$NCC_{SFOC}$ as the similarity measure.

The registration results of our system were qualitatively evaluated by visual inspection. Specifically, Fig. 16 exhibits the detected CPs of all the test cases by the designed system. We can see that these detected CPs were uniformly distributed across each multimodal image pair because the block-based FAST operator was applied to detect IPs in our system. Moreover, their location precision was satisfactory in spite of large differences in geometric distortions, NRD, noises, and spatial resolution for each multimodal image pair.

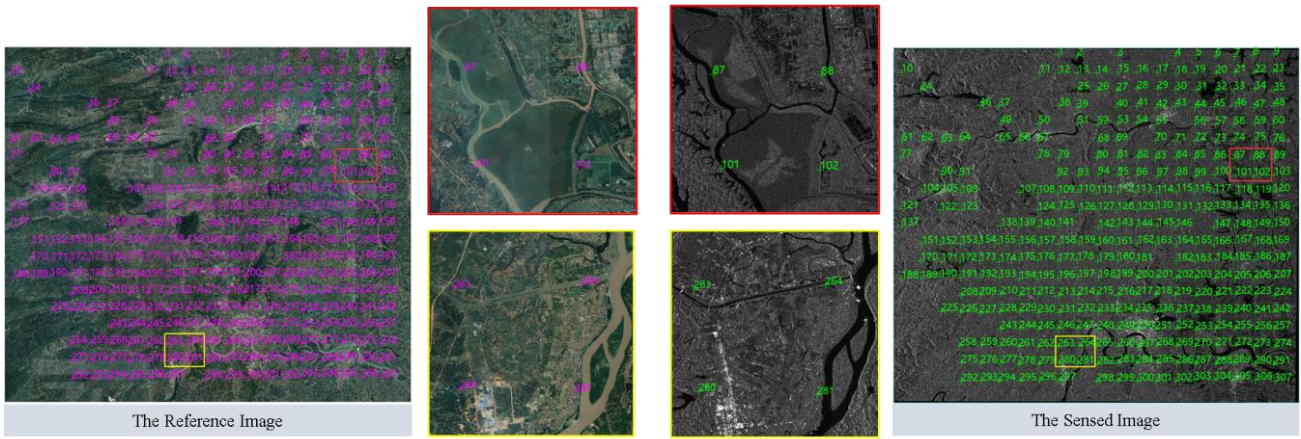

(a) Test 1

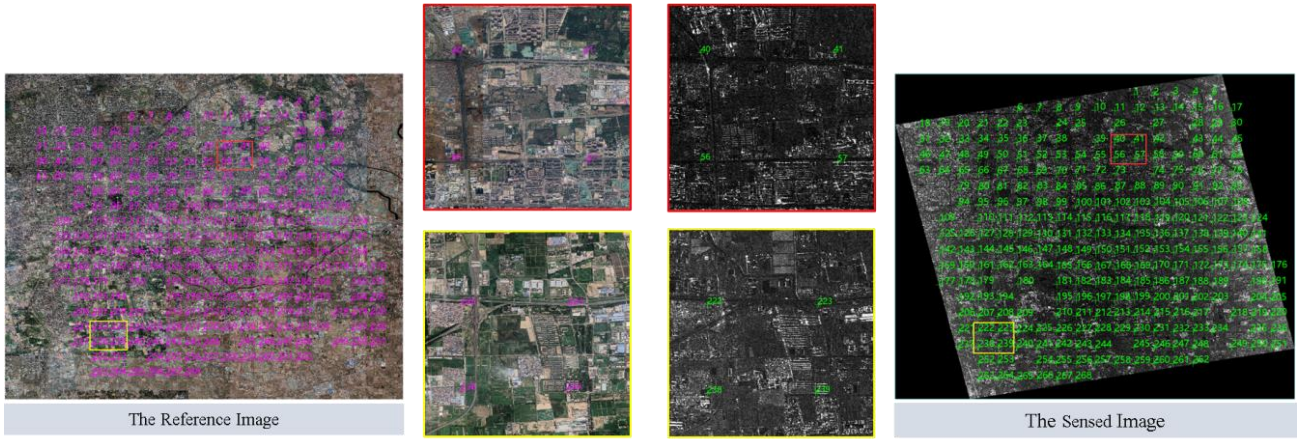

(b) Test 2

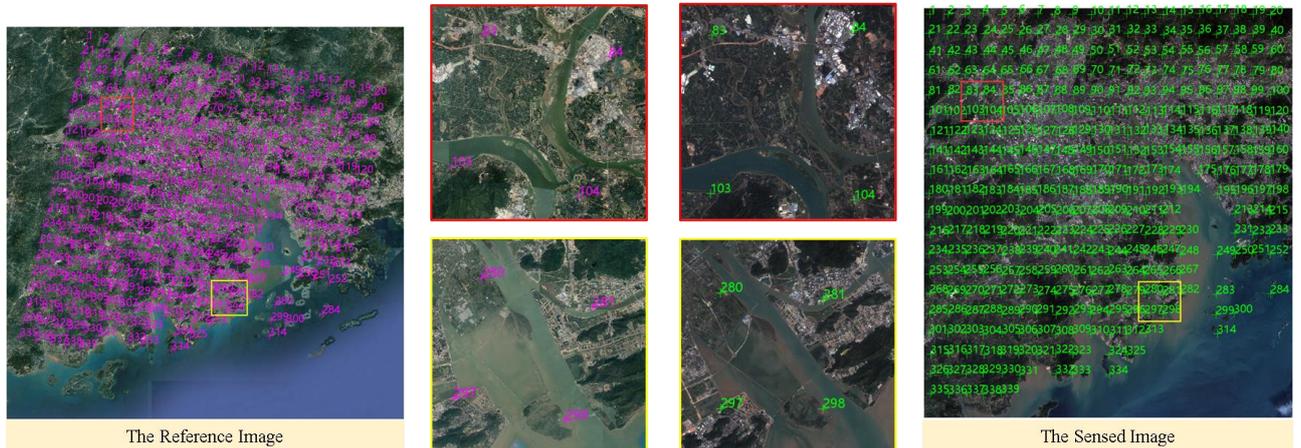

(c) Test 3

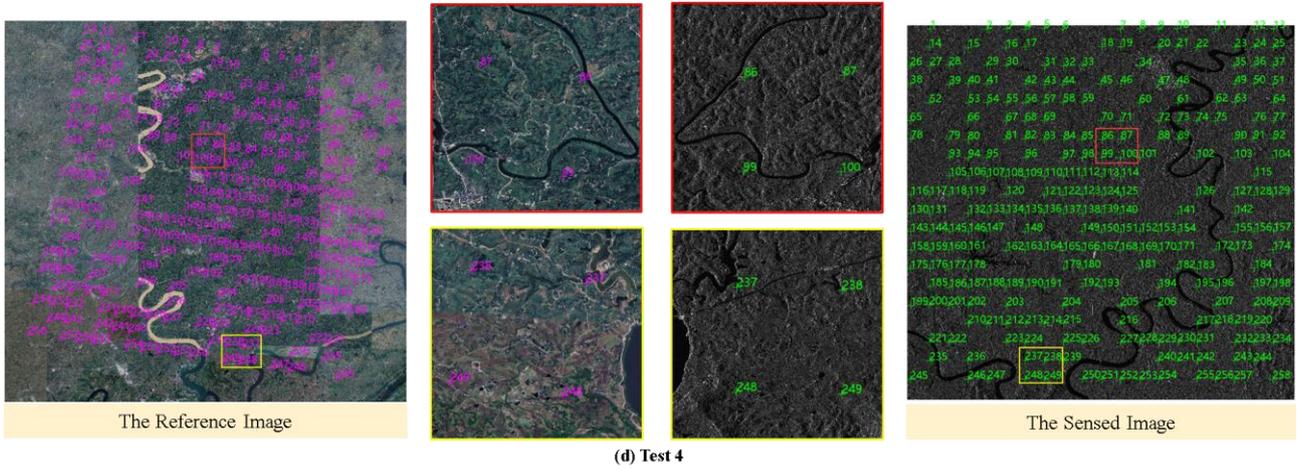

Fig. 16. Detected CPs by the designed system.

For the sake of checking the final alignment results, Fig. 17 presents the superposition registration results of the reference and sensed images registered by our system. And it was apparent that all sensed images had been aligned correctly.

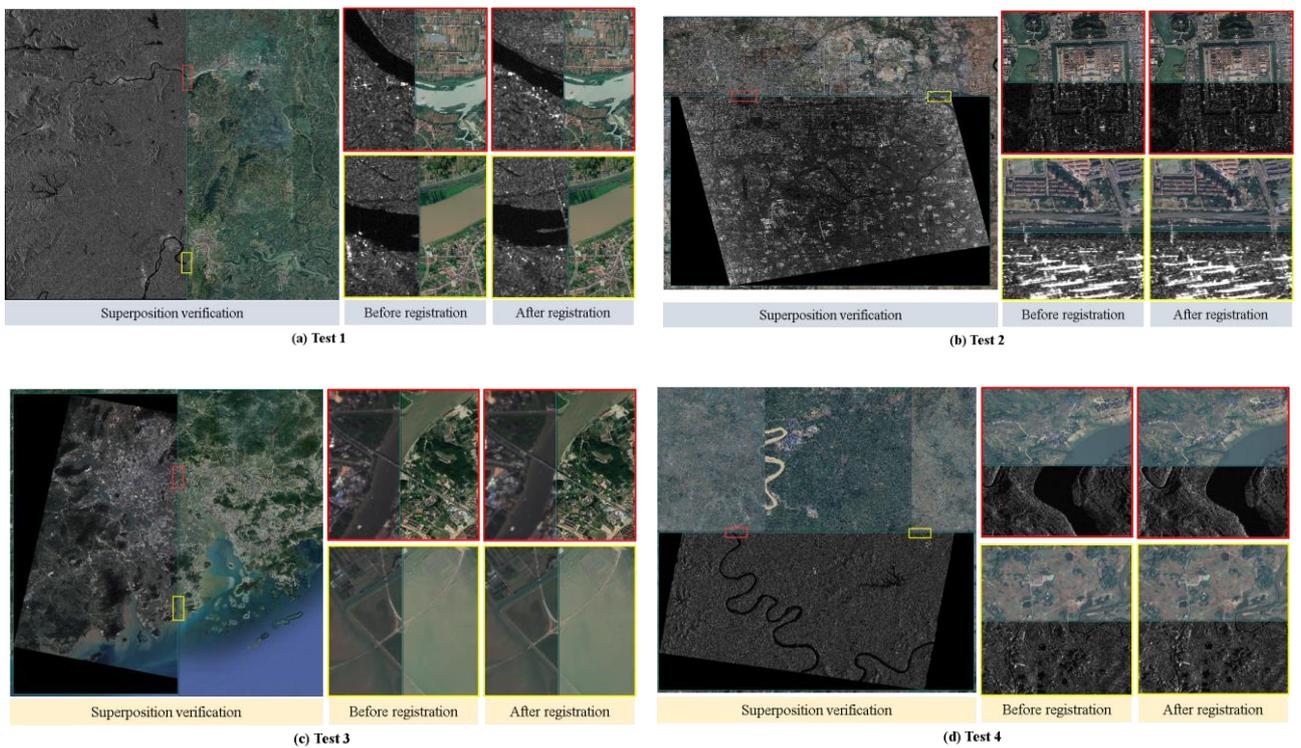

Fig. 17. Superposition registration results in terms of before registration and after registration with the designed system. (Note: the before registration of the tests 3 and 4 refer to the entire sensed image corrected by the RFM model.)

## 6. CONCLUSION

This paper has presented a robust matching method for the registration of multimodal remote sensing images, involving both a novel SFOC descriptor and a fast similarity measure (i.e., Fast-NCC$_{SFOC}$). SFOC is first proposed by making use of the first- and second-order Gaussian steerable filters, which aims to capture distinctive structural features for resisting significant NRD between multimodal images. Then Fast-NCC$_{SFOC}$ is established by combining NCC and SFOC, and it speeds up the image matching by using the FFT technique and integral images. Furthermore, an automatic registration system is developed based on the proposed matching method, which involves a local coarse registration and a fine registration. The local coarse registration is conducted using the block-based FAST operator and local geometric correction. Specifically, the block-based FAST operator is first employed to detect evenly distributed IPs, and local geometric correction is intended to eliminate the apparent geometric distortions by means of the RFM model. Next, the fine registration is employed by applying SFOG with a template matching manner. The experimental results on ten various multimodal images have demonstrated the robustness and effectiveness of SFOG. In contrast to other state-of-the-art similarity measures (i.e., the MI, MTM, PCSD, CFOG, SDFG), the proposed SFOC achieved the best matching performance both in the quantitative evaluation and qualitative examination. In addition, the designed system was also quantificationally and qualitatively evaluated by testing four multimodal images with significant differences in geometric distortions, NRD, noises, and spatial resolution. The results indicated our system outperforms ENVI, ERDAS, and PCI in registration performance, which illustrates it has the potential for engineering applications.

Although SFOC presents robust performance for multimodal image matching, it is sensitive to global geometric distortions between images, that is, it cannot be adapted to multimodal image matching with large scale or rotation differences. The designed system depends on the prior geo-referenced information of RS image to eliminate obvious geometric distortions before the fine registration. Therefore, if the sensed image does not come up with the geo-referenced information or RPCs, our system will not be applicable. Future research aims to design an enhanced system that is adaptable to geometric distortions with the assistance of geo-reference information, and explore the similarity measure with scale and rotation invariance.

## 7. ACKNOWLEDGEMENTS

The funding for the research discussed in this paper was provided by the National Natural Science Foundation of China (No.41971281). And this work was supported in part by the National Natural Science Foundation of China under Grant (No.61972021).